\documentclass[10pt,twocolumn,letterpaper]{article}

\usepackage{cvpr}      %
\usepackage{multirow}

\definecolor{cvprblue}{rgb}{0.21,0.49,0.74}
\usepackage[pagebackref,breaklinks,colorlinks,allcolors=cvprblue]{hyperref}

\def\paperID{13915} %
\def\confName{CVPR}
\def\confYear{2025}

\DeclareMathOperator*{\argmax}{arg\max}
\DeclareMathOperator*{\argmin}{arg\min}
\newcommand{\AN}[1]{ 
\textcolor[HTML]{d93b26}{[AN: #1]}
}

\newcommand{\std}[1]{\begin{tiny}{$\pm #1$}\end{tiny}}
\newcommand{\method}{\texttt{PLeaS}\ }
\newcommand{\methodw}{\texttt{PLeaS-free}\ }
\newcommand{\methodwnospace}{\texttt{PLeaS-free}}
\newcommand{\methoda}{\texttt{PLeaS-Weight}\ }
\newcommand{\methodanospace}{\texttt{PLeaS-Weight}}
\newcommand{\methodnospace}{\texttt{PLeaS}}
\title{\method --- Merging Models with Permutations and Least Squares}

\author{%
    Anshul Nasery$^\dagger$\thanks{Equal Contribution. Correspondence to anasery@cs.washington.edu} \qquad Jonathan Hayase$^\dagger$\footnotemark[1] \qquad Pang Wei Koh$^\dagger$$^\diamond$\qquad 
  Sewoong Oh$^\dagger$ \\
  \\
  \hfill $^\dagger$ University of Washington   \qquad $^\diamond$ Allen Institute for AI\hfill 
}
\begin{document}

\maketitle

\begin{abstract}
The democratization of machine learning systems has made the process of fine-tuning accessible to practitioners, leading to a wide range of open-source models fine-tuned on specialized tasks and datasets. Recent work has proposed to merge such models to combine their functionalities. 
However, prior approaches are usually restricted to models that are fine-tuned from the same base model. Furthermore, 
the final merged model is typically required to be of the same size as the original models. 
In this work, we propose a new two-step algorithm to merge models---termed \methodnospace---which relaxes these constraints.
First, leveraging the \textbf{P}ermutation symmetries inherent in the two models, \method partially matches nodes in each layer by maximizing alignment. 
Next, \method computes the weights of the merged model as a layer-wise \textbf{Lea}st \textbf{S}quares solution to minimize the approximation error between the features of the merged model and the permuted features of the original models. 
\method allows a practitioner to merge two models sharing the same architecture 
into a single performant model of a desired size, even when the two original models are fine-tuned from different base models. 
We also demonstrate how our method can be extended to address a challenging scenario where no data is available from the fine-tuning domains. 
We demonstrate our method to merge ResNet and ViT models trained with shared and different label spaces, and show improvement over the state-of-the-art merging methods  of up to 15 percentage points for the same target compute while merging models trained on DomainNet and fine-grained classification tasks\footnote{Code open-sourced at \url{https://github.com/SewoongLab/PLeaS-Merging}}.
\end{abstract}
\section{Introduction}

With the widespread democratization of machine learning, there has been a rapid increase in the availability of open-source models trained by the community on specific tasks and datasets. Such specialized models exhibit unique strengths and weaknesses. For example, Code Llama \citep{roziere2023code} (fine-tuned from Llama-2) is specialized for coding, while Vicuña 1.3 \citep{vicuna2023} (fine-tuned from Llama-1) is specialized for chat. They have the same architecture but 
are fine-tuned starting from different pre-trained models: Llama-1 and Llama-2. 
Such diversity in the combination of pre-training data and fine-tuning tasks will only increase as decentralized marketplaces for models become increasingly more common, \eg, \cite{rao2020bittensor}, providing practitioners with more options.

This presents an opportunity to combine such specialized models in order to create a single general-purpose model that can handle multiple tasks.  
Traditional approaches for combining trained models, such as ensembling \cite{Ganaie_2022} or domain-specific mixture-of-experts (\eg\cite{jain2023damex}), take a step towards this goal. However, these methods need to store all the component models at inference time, leading to an increased memory footprint. Practitioners with limited memory capacity cannot use such approaches with high and fixed memory costs,   
especially when combining large models, deploying to resource-constrained environments, or for applications demanding a memory-performance trade-off. 

To this end, recent works \cite{yadav2023tiesmerging, ilharco2023editing, yang2024adamerging, yang2024representation} have proposed new algorithms tackling this problem of \textit{model merging}. However, their scope is limited to merging models fine-tuned from the \textit{same} pretrained model. Further, some recent works \cite{stoica2024zipit} also need access to the \textit{training data} used to fine-tune the component models, which limits their applicability in situations where such data is not available due to, for example, privacy or legal reasons \cite{demircan2022dma}. 
In this paper, we address the problem of merging models (sharing the same architecture) trained on different datasets starting from \textit{different initializations}.
This is motivated by prior work (\eg, \cite{stoica2024zipit, yamada2023revisiting, NEURIPS2018_ad8e88c0}), which we compare with in Section~\ref{sec:experiments} for merging ResNet and ViT models. %
To address the above limitations of prior work in this space, we present \methodnospace---an algorithm which adaptively merges models for different inference compute budgets, and can work without using the fine-tuning data of the component models.
\begin{figure*}[t]
  \centering
  {\includegraphics[width=0.9\linewidth]{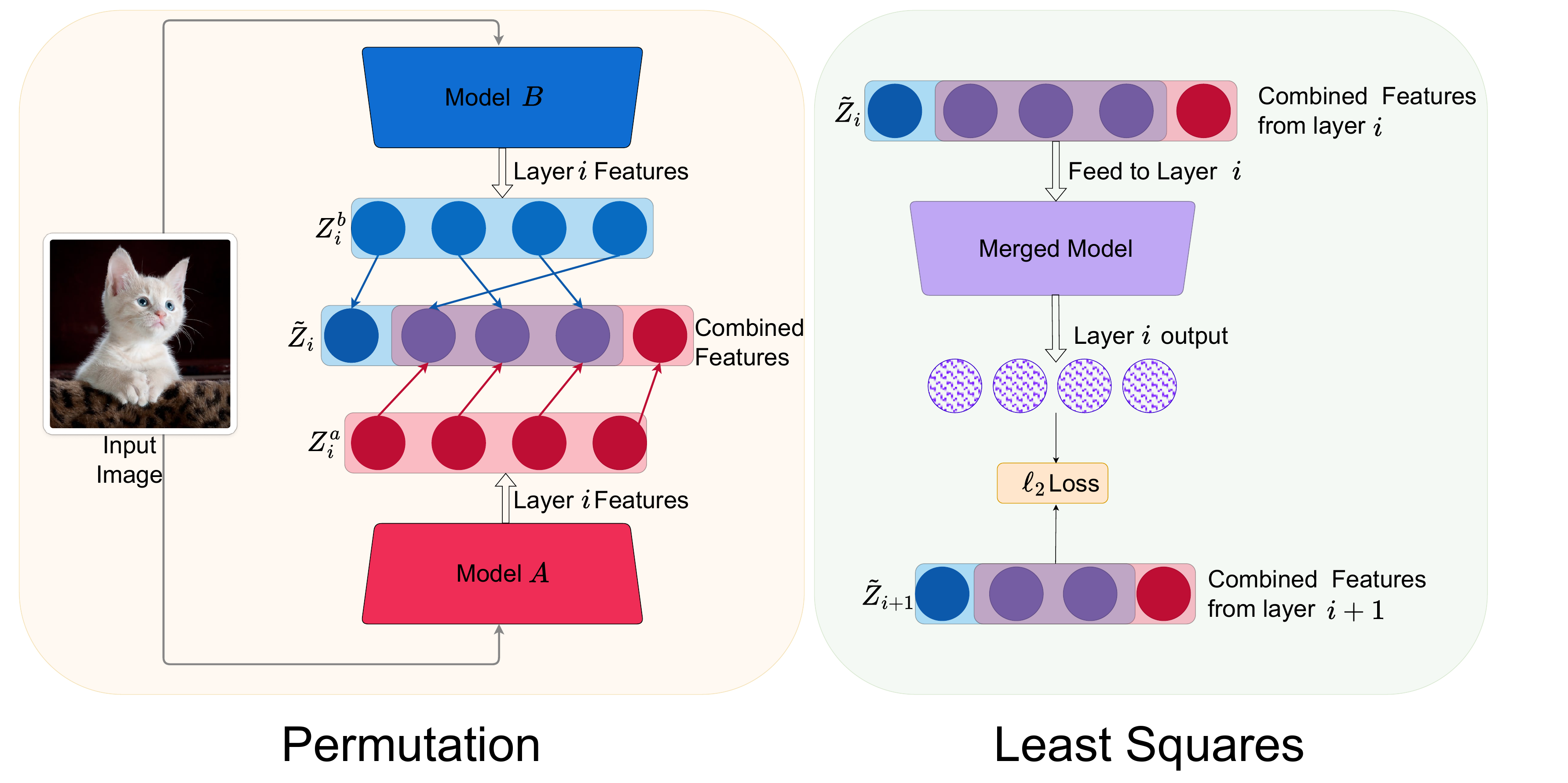}}
  \caption{\textbf{\method is a two-step algorithm for merging models}: %
  The first step (left) finds layer-wise \textbf{P}ermutations to match features across models to compute combined features $\tilde{Z_i}$. Features which are similar are \textcolor[HTML]{735ca1}{merged}, while those which are dis-similar are \textcolor[HTML]{BE0E34}{kept} \textcolor[HTML]{086BC2}{separate}. The number of features to be merged depends on the target compute budget, and can be different for each layer. The second step of \method (right) aims to find weights of the merged model which can map the combined features of layer $i$ (i.e., $\tilde{Z}_i$) to those of layer $i+1$ (i.e., $\tilde{Z}_{i+1}$) appropriately by solving layer-wise \textbf{Lea}st \textbf{S}quares problems for each layer.
  } 
  \label{fig:algo}
\end{figure*}

\method (short for \textbf{P}ermutations and \textbf{Lea}st \textbf{S}quares) is a two-stage algorithm which works with models having the same architecture. 
The first step consists of matching features across the models. We harness the idea of permutation invariance in neural networks to find an appropriate pairing of features. Inspired by the Git Re-Basin \cite{ainsworth2022git} algorithm, which is designed for merging two models that are trained on the same data, we introduce a matching algorithm that finds permutations between similar features across models, while keeping dissimilar features separate in the final merged model. This is critical when merging models trained on widely different tasks, since it prevents interference between features while still merging overlapping features.
This also gives \method fine-grained control over the width of each layer of the merged model, improving performance over prior work such as ZipIt!~\cite{stoica2024zipit}. \method{} can hence flexibly trade-off inference memory/compute and performance according to the deployment requirements.

It has been observed that permutation matching alone suffers from significant performance loss when merging vastly different models, \eg those trained on disparate data \cite{yamada2023revisiting}. We hypothesize that while permuted features are powerful when ensembled, simply averaging the permuted weights degrades the features of the merged model. This results in the observed decline in performance. Hence, in the second step of \methodnospace, we solve a layer-wise Least Squares problem, so that each layer of the merged model mimics the permuted ensemble of features from the corresponding layer of the original models. This produces better representations and superior down-stream performance. 

Apart from the target compute budget, \method is hyperparameter free, making it easy for practitioners to use. 
A schematic of \method is depicted in \cref{fig:algo}. 

We empirically demonstrate that \method can outperform prior work in the challenging setting of merging differently initialized models which have been trained on different datasets.
We merge ResNet and ViT models fine-tuned on different datasets in \cref{sec:same_size,sec:diff_size,sec:same_initialization}, and find that \method improves upon the state-of-the-art up to 15\% with the same merged model size. Our empirical results are on subsets of DomainNet, and on fine-grained classification tasks. {\methodnospace} can also approach the performance of ensemble methods with significantly lower FLOPs (\cref{sec:same_initialization}).

The proposed approach can be seamlessly extended to the scenario where data from the fine-tuning domains is unavailable. We 
call this variant \methodwnospace. This variant uses data from publicly available datasets (like ImageNet) to merge models.
We demonstrate in \cref{sec:no_data} that \methodw is competitive with \methodnospace, which uses the actual data from the training domains of the component models. 
This is highly encouraging, as it demonstrates the applicability of \methodw in scenarios where data from the training domains is unavailable due to privacy or commercial reasons.  

In summary, our contributions are the following:
\begin{itemize}
    \item We generalize Git Re-Basin \cite{ainsworth2022git} to support partial merging of corresponding  layers of two models (\cref{sec:partial-merge}). 
     This gives practitioners the freedom to choose the size of the final merged model as per resources available at inference. Investigating this tradeoff is one of the goals in this work, \eg, \cref{fig:mem_performance_tradeoff}. 
    \item  Motivated by the success of ensemble methods, we propose to assign weights to the merged model's parameters by solving a least squares problem attempting to mimic ensemble methods at each layer  (\cref{sec:step2}). Ablation study for this step is in \cref{fig:mem_performance_tradeoff}.
    \item On a test-bed of multiple datasets, 
     we showcase that \method     outperforms recent merging methods up to 15 percentage points (\cref{sec:experiments}) at the same model size. Further, \method approaches the ensemble accuracy while using  40\% fewer parameters. Finally, even with no data from the training domains, \methodw remains competitive with \method  (\cref{fig:main_figure4}). \end{itemize}
\section{Related works}

There has been growing interest in merging models with minimal data and compute overhead. 
Here, we focus on methods which merge models with the \textit{same} architecture.

\textbf{Merging models fine-tuned from the same initialization.} Several methods aim to merge models in the weight space. \citet{ilharco2023editing} simply add up {\em task vectors}, the weight differences of fine-tuned models from the pretrained model, and demonstrate a strong baseline for merging fine-tuned models. Other approaches edit the task vectors based on magnitude of the weights \cite{yadav2023tiesmerging, yu2024language} to resolve interference while merging. Some methods aim to find layer-wise \cite{yang2024adamerging, akiba2024evolutionary} or parameter-wise \cite{matena2022merging} coefficients for merging different task vectors.
However, methods that work with task vectors assume that the base pretrained model is shared across the fine-tuned models, and hence they cannot be easily extended to settings where models are fine-tuned from different starting points. 
A different line of work \cite{jin2023dataless} proposes layer wise distillation, aiming to minimize the sum of the $\ell_2$ distances between the activations of the merged model and the original models. However, naively applying this to vastly different models leads to degraded performance, as we show in \cref{sec:experiments}. Further, these methods do not provide a way to control the size of the merged model. Although not designed for this scenario of merging fine-tunes of a common pre-trained model, \method still allows us to achieve significant performance gains when the merged network is slightly larger than the original model (\eg by 20\%)  as demonstrated in \cref{tab:same_init}. 

\textbf{Merging from two different initializations.}
We consider a less restrictive setting, where the models being merged can have different initializations. This has been studied in the literature, and existing works propose weight or activation matching algorithms for this task. Git Re-Basin~\cite{ainsworth2022git} proposes an algorithm to compute permutation matrices to match the weights of the hidden layers of two or more neural networks. \citet{yamada2023revisiting} investigate the usage of permutations to merge models trained on different datasets, however, their study is limited to wide ResNet models on MNIST and CIFAR datasets. 
These permutation symmetries have also been studied in \cite{navon2023equivariant, yamada2023revisiting, cho2024sherpa,singh2020model}. A recent work -- MuDSC \cite{xu2024training} leverages permutation symmetries both in weight space and activation space to merge models better, however, we show that \method{} outperforms this work empirically.  
Another recent work -- ZipIt!~\cite{stoica2024zipit}, tackles a similar problem of merging models fine-tuned on different datasets from different initializations.
This work also supports merging models partially by ``zipping'' some layers of the component models. While this can provide a knob for controlling the size-performance trade-off of the merged model, the empirical performance of their proposed scheme can be improved upon, as we show in \cref{sec:size-accuracy-tradeoff}. On the other hand, our work describes a merging formulation which is more expressive and allows for partial merges with expanded layers to minimize feature interference. 
Finally, \cite{NEURIPS2018_ad8e88c0} also proposes a method to merge networks layer-wise in a progressive manner, which involves light-weight retraining. 
However, their method requires domain labeled data at both training and inference time, while we only require unlabeled data and also propose a method using no data from the train domain at all.

\textbf{Other merging paradigms.}
Other model merging approaches include mixture of experts \citep{shazeer2017outrageously,sukhbaatar2024branch}, selecting experts using test data \citep{li2022branch}, and sparse expert ensembles \citep{gururangan2023scaling}. These come with larger compute or memory over-heads, both at inference and training time.

\section{Preliminaries}
{\bf Notation.} 
For simplicity, we describe our method for two $L$-layered MLPs. However, it can be readily extended to convolutional and residual networks, as we demonstrate in our experiments. Let $\Theta^A = \{W^A_1,W^A_2,\cdots,W^A_L\}, \Theta^B=\{W^B_1,W^B_2,\cdots,W^B_L\}$ be the parameters of two MLPs $A,B$ having the same architecture. We omit the layer-wise bias here for simplicity. Let $z_i^A, z_i^B$ denote the input activations to the $i$\textsuperscript{th} layer of each network respectively, and $d_i$ denote the dimension of $z_i^A,z_i^B$. We also define $Z_i^A, Z_i^B \in \mathbb{R}^{d_i \times n}$ to be the activations of a batch of $n$ inputs. Note that $z_1^A=z_1^B=x$, and $z_{L+1}^A = y^A, z_{L+1}^B = y^B$. Finally, let $\{W_i^M:i \in \{1,2,\cdots,L\}\}$ be the weights of the merged model. We allow the merged model to have varying widths (which can be different from the widths of the base model), depending on the inference resources available.

{\bf Background on Git Re-Basin.} 
Our method is inspired by  Git Re-Basin \cite{ainsworth2022git}, which aims to find permutation matrices $\pi = \{P_1, P_2, \cdots, P_L\}$ to permute the weights of model $B$.
The merged model is formed by permuting and averaging the weights, i.e., $W_i^M = (1/2) (W_i^A + P_i W_i^B P_{i-1}^T)$.  

\citet{ainsworth2022git} propose to estimate the permutation matrices by directly optimizing the average similarity between the permuted weights of model $B$ and the original weights of model $A$. This \textit{weight matching} greedily finds a solution to the following sum of bilinear assignment problems, 
\begin{equation*}
    \argmax_{\pi=\{P_i\}_{i=1}^L} \;\; \sum_{i=1}^L \;\langle W_i^A, P_i W_i^B P_{i-1}^T \rangle\;, 
\end{equation*}
where $P_{0}$ is defined to be the identity matrix. This has an advantage of not requiring any data to solve the optimization, but an optimal solution is computationally intractable. 
Instead, when some samples are available to the optimizer, \citet{ainsworth2022git} propose a computationally efficient alternative called \textit{activation matching}, which solves the following optimization problem: 
\begin{equation*}
    P_i \; \in \;  \argmin_{P \in S_{d_i}} \; \|Z_{i}^A - PZ_{i}^B \|_F^2 \;. 
\end{equation*} 
Here, $S_{d_i}$ refers to the set of permutation matrices of size $d_i \times d_i$. 
Computing the activations $Z$'s require samples from the data. However, this optimization can be efficiently solved separately for each layer. 

\section{Method: \method}
We call our approach to model merging \methodnospace. 
We harness permutation symmetries to match features between two models, inspired by Git Re-Basin \cite{ainsworth2022git}. We extend this method to allow for partial merging of models, where each layer can have a different number of merged neurons. 

We then compute the weights of the final merged model by solving layer wise least squares problems to ensure that activations of the merged model resemble the permuted activations of the original models.

\subsection{Extending Git Re-Basin to partial merging}\label{sec:partial-merge}  

Note that in Git Re-Basin, two models are averaged (after permuting one model) and hence the dimension of the parameters of the merged model is the same as the corresponding parameters of the base models. However, when the networks $A,B$ are trained on different datasets, not all features might be compatible across models.
These features may interfere destructively if merged,
leading to degraded performance. Further, these incompatible features may need to be retained in the merged model in order to make accurate predictions on both tasks. 
Merging all nodes in every layer discounts this possibility, leading to performance degradation, as we show in \cref{fig:data_diff_label_spaces}.
To this end, we aim to merge features which are similar across the two models, while keeping those which are very different as separate features in the merged model. We hence propose a framework for partially merging model features by leveraging permutations. 

Given a permutation matrix $P_i$, we select $k_i$ indices from $[d_i]$ for which the distance between the features of model $A$ and the permuted features of model $B$ for layer $i$ is the smallest. These $k_i$ features are merged, while other features are retained separately in the final model. 
\begin{figure}
\centering
\includegraphics[width=0.8\linewidth]{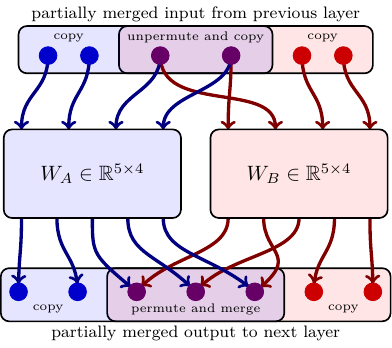}
\caption{\textbf{Partial merging with permutations}: We show the construction of the \(7 \times 6\) weight matrix \(W_i^m\) from two weights of size \(5 \times 4\) in the first step of \methodnospace. The merged inputs are copied and unpermuted to approximate the original inputs. Then we apply both weight matrices separately. Finally, we pair up the merged outputs and average the pairs. Since all operations used are linear, we can fuse them to construct \(W_i^m\) using a single linear layer.}
\label{fig:perm_method}
\end{figure}
In particular, we find a subset $J_i$ satisfying
\[J_i \in  \argmin_{\{J : J\subseteq[d_i], |J|=k_i\}}  ||Z_{J,i}^A - (P_iZ_{:,i}^B)_J ||_F^2\;.\]  
This is simple to implement: retain the indices with the smallest $k_i$ distances between the (permuted) activations. For weight matching, we can retain the indices with the largest similarity between the (permuted) weights for each layer. 
The size of $W_i^M$ is then increased to $(2d_i-k_i) \times (2d_{i+1}-k_{i+1})$ in exchange for improved performance.
This partial merging scheme is illustrated in \cref{fig:perm_method}.
One goal of this paper is to investigate this trade-off between size and the performance of the merged model. 

Note that the ratio ${k_i}/{d_i}$ can be chosen independently for each layer. In \cref{app:qp}, we propose a scheme to find a configuration of these ratios subject to a target compute/memory budget $B$; this optimizes a proxy of the downstream performance without using any validation data from the target domain, and is used in all our experiments. 
The permutation matrices $P_i$ are computed using the activation matching strategy from Git Re-Basin. In \cref{sec:no_data} and the Appendix, we compare this with using the weight matching strategy, which we call \methoda. 

We would like to emphasize that our partial merging formulation is different from ZipIt!~\cite{stoica2024zipit}, since we can assign any amount of compute between \(1\times\) and \(2\times\) (relative to the original layer's compute) to each layer independently.
ZipIt!\ on the other hand, assigns exactly \(1\times\) compute to a prefix of layers and \(2\times\) to the rest. In \cref{sec:diff_size}, we show that this flexibility in our formulation leads to better empirical performance.

\subsection{Permuted least squares} \label{sec:step2}

Suppose, for example, that the target merged model has the same architecture and size as each of the base models. Once the permutations, $P_i$, have been computed, we propose optimizing the weight matrices of the merged model by solving the following least-squares problem: 
\begin{equation}
\label{eqn:perm_reg}
\resizebox{\linewidth}{!}{$
    W_i^{M} \in \smash{\argmin\limits_{W}} \|(Z_{i}^A + P_i Z_{i}^B) W - (Z_{i+1}^A + P_{i+1} Z_{i+1}^B) \|^2,
$}\end{equation}
independently for each layer $i\in[L]$. This is motivated by the impressive performance of the ensemble method (\eg, \cite{yamada2023revisiting} and \cref{sec:experiments}), which retains two separate models and only averages the (permuted) activations at the last layer (pre-softmax): $\tilde{z}_{L+1} = z^A_{L+1}+z^B_{L+1}$. 
We aim to have our merged model approximate such activations. 
We inductively assume that the first $i-1$ layers are properly merged. Hence, the ensemble of the permuted features (of the $i$\textsuperscript{th} layer) of the component models can be well approximated by the activations at the input of the $i$\textsuperscript{th} layer of the merged model. We denote the ensembled features by $\tilde{Z}_{i}=(Z_{i}^A+P_iZ_{i}^B) \in {\mathbb R}^{d_i\times n}$. The goal of the above optimization is to match the ensembled activation of the next layer, $\tilde{Z}_{i+1}=(Z_{i+1}^A+P_{i+1}Z_{i+1}^B)$, with a linear transform of the input ensemble: $\tilde{Z}_{i} W$. We empirically validate this choice to use a permuted ensemble of features to optimize the weights of the merged model in \cref{tab:method_ablation} in \cref{app:ablations}, where we compare with alternatives to \cref{eqn:perm_reg}.  

This second step of \method is similar to feature distillation. However, the key novelty arises from averaging the permuted features for transferring knowledge from multiple models. This is critical for accurate prediction. To show this, we compare \method against RegMean~\cite{jin2023dataless}, which optimizes an objective similar to \cref{eqn:perm_reg} without the permutations and averaging, i.e. this method merges models by minimizing $\|Z_{i}^A W - Z_{i+1}^A \|^2 + \|Z_{i}^B W - Z_{i+1}^B \|^2$. As we show in \cref{sec:experiments}, RegMean performs poorly compared to \methodnospace. 
Apart from the inference computation budget for the final model, \method is completely hyperparameter free.

Note that the second step of \method is fully compatible with the partial merging of \cref{sec:partial-merge}
as well: we can directly set the values of $W_i^M$ corresponding to the unmerged features to be the respective values of $W_i^A$ and $W_i^B$.

While the objective in \cref{eqn:perm_reg} can be minimized in closed form using Ordinary Least Squares (OLS), we practically implement it using gradient descent for ease of use with convolution layers. Given that the objective is convex if computed layer-wise, the weight matrices $W_i^M$ converge in relatively few (less than 100) steps of gradient descent. Further, we solve this optimization independently for each layer, so it can be efficiently parallelized.

\subsection{Data requirements of \method}
\method has two steps -- the first step finds permutations to match features using weight or activation matching and the second step computes weight matrices to mimic the ensemble of the merged features more closely. In order to compute these features, one could use the data from the training domains, however, this may not be feasible for privacy or commercial reasons. Hence, we propose an alternative scheme--dubbed \methodwnospace--which uses a general vision dataset, like ImageNet, to compute the activations of the component models. These activations are then used to merge domain specific models without requiring any data from the training domains. In \cref{sec:no_data}, we show that \methodw suffers a minimal performance penalty compared to \methodnospace, suggesting wider applicability in low/no data settings.  
 
\section{Experiments}
\def\oracleTable#1{\begin{table*}[]
\centering
\caption{\textbf{Evaluating merged models with task specific heads}: For each pair of datasets, we merge models and compute the final accuracy on the pair. To compute the final accuracy on a dataset, we average the accuracies across the pairs that it is a part of. These accuracies are computed by using the task specific head for each pair of models. }
\label{tab:oracle_results}
\begin{tabular}{llllll}
\toprule
Method & FLOPs & CUB & NABird & Dogs & Pets     \\
\midrule
ZipIt\cite{stoica2024zipit} & 1.0 & 47.8 & 46.3 & 35.7 & 68.6 \\
Git Re-Basin\cite{ainsworth2022git} & 1.0 & 35.3 & 32.7 & 26.1 & 50.4   \\
\method & 1.0 & 63.1 & 63.2 & 45.3 & 68.4  \\
\midrule
ZipIt & 1.2 &  68.7 & 64.1 & 60.9 & 84.8 \\
Permutation & 1.2 & 67.8 & 64.7 & 60.9 & 83.7  \\
\method & 1.2 & 71.6 & 69.6 & 70.2 & 86.9 \\
\midrule
ZipIt & 1.55 & 72.9 & 69.2 & 71.9 & 89.7 \\
Permutation & 1.55 & 71.8 & 69.8 & 71.8 & 88.2 \\
\method & 1.55 & 74.5 & 72.4 & 74.8 & 89.8\\
\midrule
ZipIt & 1.8 & 75.1 & 72.7 & 74.0 & 91.6 \\
Permutation & 1.8 & 74.1 & 72.6 & 75.8 & 89.9  \\
\method & 1.8 & 77.0 & 74.3 & 75.5 & 89.6 \\
\midrule
Ensemble & 2.0 &  77.0 & 76.0 & 80.8 & 91.6  \\
\bottomrule
\end{tabular}
\end{table*}}

\def\mainTableAverageSingleBudget#1{\begin{table*}[]
\centering
\begin{tabular}{llllllllllllll}
\toprule
\multirow{2.3}{*}{Method} & \multirow{2.3}{*}{FLOPs} & \multirow{2.3}{*}{Memory} & \multicolumn{5}{c}{Same Label Space} & \multicolumn{5}{c}{Different Label Spaces} \\
\cmidrule(lr){4-8} \cmidrule(lr){9-13}
       &       &        & Clip & Info & Paint & Real & Avg & CUB & Pets & Dogs & NABird & Avg \\
\midrule
\textcolor{gray}{MoE*}   & \textcolor{gray}{$1.1\times$} & \textcolor{gray}{$2.1\times$} & \textcolor{gray}{69.1} & \textcolor{gray}{36.1} & \textcolor{gray}{65.7} & \textcolor{gray}{78.0} & \textcolor{gray}{62.2} & \textcolor{gray}{81.1} & \textcolor{gray}{92.7} & \textcolor{gray}{83.1} & \textcolor{gray}{75.8} & \textcolor{gray}{83.2} \\
\textcolor{gray}{Ensemble*} & \textcolor{gray}{$2\times$} & \textcolor{gray}{$2\times$} & \textcolor{gray}{63.6} & \textcolor{gray}{30.3} & \textcolor{gray}{61.0} & \textcolor{gray}{74.7} & \textcolor{gray}{57.4} & \textcolor{gray}{80.5} & \textcolor{gray}{92.8} & \textcolor{gray}{82.1} & \textcolor{gray}{76.1} & \textcolor{gray}{82.9} \\
\midrule
Simple Avg~\cite{ilharco2023editing} & $1\times$ & $1\times$ & 1.2 & 0.8 & 1.9 & 2.1 & 1.5 & 7.1 & 19.2 & 9.2 & 4.7 & 10.1 \\
RegMean~\cite{jin2023dataless} & $1\times$ & $1\times$ & 16.6 & 5.8 & 10.1 & 15.8 & 12.1 & 42.5 & 45.1 & 20.2 & 37.1 & 36.2 \\
ZipIt!~\cite{stoica2024zipit} & $1\times$ & $1\times$ & 26.9 & 12.2 & 27.1 & 37.4 & 25.9 & 67.5 & 83.6 & 60.0 & 56.3 & 66.9 \\
Git Re-Basin~\cite{ainsworth2022git} & $1\times$ & $1\times$ & 18.2 & 7.8 & 18.8 & 26.5 & 17.8 & 66.2 & 80.2 & 62.6 & 59.4 & 67.1 \\
MuDSC~\cite{xu2024training} & $1\times$  & $1\times$ & 34.0 & 14.3 & 29.5 & 45.1 & 30.7 & 70.1 & 82.5 & 63.2 & 58.2 & 68.5 \\
\method (Ours) & $1\times$ & $1\times$ & \textbf{41.7} & \textbf{16.9} & \textbf{40.8} & \textbf{55.1} & \textbf{38.6} & \textbf{75.2} & \textbf{85.0} & \textbf{69.6} & \textbf{69.7} & \textbf{74.9} \\
\bottomrule
\end{tabular}
\caption{
\textbf{Merging pairs of models trained on different datasets}: For each pair of datasets, we merge models and compute the final accuracy on the pair. To compute the final accuracy for a dataset, we average the accuracies across the pairs that the dataset is a part of. We report the accuracies of the merged models for the Same Label Space setting, and a linear probe accuracy on the representations of the merged model for the Different Label Space setting. * Note that here the  merged models (bottom six) have the same size as the original, but the  MoE and ensemble have a size twice the original. 
}

\label{tab:main_results} 

\end{table*}
}

\def\methodAblations#1{\begin{table*}[]
\centering
\caption{\textbf{Comparing different objectives for \method}: We compare the performance of different loss functions for the Least Squares component of \method. We find that \cref{eqn:perm_reg} gives the best performance on DomainNet and ImageNet when merging models completely. Here $\tilde{Z_i}=0.5(Z_{:,i}^a+P_{i}Z_{:,i}^b)$ and $\tilde{Z}_{i+1} = 0.5(Z_{:,i+1}^a+P_{i+1}Z_{:,i+1}^b)$}
\label{tab:method_ablation}
\begin{tabular}{lll}
\toprule
Optimization Objective & DomainNet & ImageNet \\
\midrule
$\begin{aligned}
&\|Z_{:,i}^a W - \tilde{Z}_{i+1} \|^2 +
\|P_{i}Z_{:,i}^b W - \tilde{Z}_{i+1} \|^2
\end{aligned}$              & 22.3      & 45.1     \\
$\begin{aligned}
&\|Z_{:,i}^a W - (Z_{:,i+1}^a) \|^2 + 
\|P_{i}Z_{:,i}^b W - (P_{i+1}Z_{:,i+1}^b) \|^2
\end{aligned}$      & 30.6      & 53.2     \\
$\begin{aligned}
&\|\tilde{Z_i} W - Z_{:,i+1}^a \|^2 + 
\|\tilde{Z_i} W - P_{i+1}Z_{:,i+1}^b \|^2
\end{aligned}$               & 34.3      & 58.1     \\
$\|\tilde{Z_i} W - \tilde{Z}_{i+1} \|^2$& \textbf{40.1}      & \textbf{63.1}     \\
\bottomrule
\end{tabular}
\end{table*}}

\def\sameInitTable#1{\begin{table}[]
    \centering
    \resizebox{\linewidth}{!}{
    \begin{tabular}{llllll}
    \toprule
    Method         & Size & Clipart & Infograph & Painting & Real \\
    \midrule
      Simple Avg & 1.0 & 58.2 & 28.9 & 55.7 & 70.2    \\
      RegMean &  1.0  & 58.9 & 29.0 & 57.4 & 71.8   \\
      MuDSC & 1.0 & 57.8 & 55.8 & 55.1 & 67.6 \\
      \method & 1.0 & 59.6 & 29.5 & 58.0 & 72.0  \\
    \midrule
      \method & 1.2 & 64.2 & 31.9 & 61.8 & 75.9   \\
    \midrule
    Ensemble & 2.0 & 64.4 & 32.0 & 62.0 & 76.1 \\
    \bottomrule
    \end{tabular}
    }
        \caption{\textbf{Merging models fine-tuned from the same initialization}: We merge pairs of models fine-tuned from the same base model and compute the average accuracy across all pairs of domains for each dataset in the Shared Label Setting. We find that \method can approach the performance of the ensemble while having a $1.2\times$ sized merged model.}
    \vspace{1ex}
    \label{tab:same_init}

\end{table}}

\def\mainTableSingleBudget#1{\begin{table}[]
\resizebox{\textwidth}{!}{
\begin{tabular}{lllllllllllllllllllllllll}
\toprule
Method & \multicolumn{12}{c}{Same Label Space} & \multicolumn{12}{c}{Different Label Space} \\
                & \multicolumn{2}{c}{Cl-Pa} & \multicolumn{2}{c}{Cl-Re} & \multicolumn{2}{c}{Cl-In} & \multicolumn{2}{c}{Pa-Re} & \multicolumn{2}{c}{Pa-In} & \multicolumn{2}{c}{Re-In} & \multicolumn{2}{c}{Cu-Na} & \multicolumn{2}{l}{Cu-Ox} & \multicolumn{2}{c}{Cu-St} & \multicolumn{2}{c}{Na-St} & \multicolumn{2}{c}{Na-Ox} & \multicolumn{2}{c}{St-Ox} \\
\midrule
MoE                    &             &             &             &             &             &             &             &             &             &             &             &             &             &             &             &             &             &             &             &             &             &             &             &             \\
Ensemble*              &             &             &             &             &             &             &             &             &             &             &             &             &             &             &             &             &             &             &             &             &             &             &             &             \\
\midrule
SimpleAvg              &             &             &             &             &             &             &             &             &             &             &             &             &             &             &             &             &             &             &             &             &             &             &             &             \\
RegMean & 18.5 \std{3.6} & 6.7 \std{0.5} & 13.3 \std{0.1} & 9.8 \std{1.9} & 18.1 \std{3.3} & 20.4 \std{5.5} & 12.2 \std{3.7} & 5.7 \std{2.4} & 8.5 \std{4.4} & 11.8 \std{5.6} & 4.9 \std{0.1} & 15.2 \std{1.7} &            58.7 & 44.8 & 62.3 & 58.4 & 57.8 & 36.8 & 58.4 & 38.5 & nan & nan & 57.4 & 73.7       \\
ZipIt                  &   51.0 \std{0.3} & 25.0 \std{0.2} & 54.2 \std{0.3} & 50.4 \std{0.3} & 58.4 \std{1.4} & 66.5 \std{0.7} & 47.4 \std{0.6} & 23.9 \std{0.2} & 56.9 \std{0.9} & 68.0 \std{1.0} & 65.0 \std{0.0} & 27.4 \std{0.0} &            72.1 & 57.3 & 66.2 & 80.5 & 64.5 & 58.2 & nan & nan & nan & nan & nan & nan            \\
GRB                    &             &             &             &             &             &             &             &             &             &             &             &             &            73.4 & 60.7 & 67.5 & 79.6 & 60.1 & 58.3 & 56.5 & 54.8 & 61.4 & 75.6 & 72.3 & 86.6        \\
\method & 55.9\std{0.6} & 27.6\std{0.2} & 56.7\std{1.0} & 54.6\std{0.3} & 59.9\std{0.4} & 70.3\std{0.7} & 53.8\std{0.2} & 26.7\std{0.8} & 59.1\std{0.4} & 71.3\std{0.5} & 28.1\std{0.3} & 70.4\std{0.6} &            79.8 & 70.0 & 74.8 & 86.9 & 71.3 & 69.7 & 68.6 & 61.8 & nan & nan & 78.1 & 90.7 \\            
\bottomrule
\end{tabular}}

\end{table}}

\def\tableOracle#1{\begin{table}[]
\centering
\caption{\textbf{Merging models trained on different datasets}: For each pair of datasets, we merge models and compute the accuracy on the pair. To compute the final accuracy on a dataset, we average the accuracies across the pairs that it is a part of.\AN{Add memory footprints, shift caption to above table} 
}
\label{tab:oracle_results}
\resizebox{\textwidth}{!}{%
\begin{tabular}{llllllllllll}
\toprule
Method & FLOPs & \multicolumn{5}{c}{Same Label Space} & \multicolumn{5}{c}{Different Label Spaces} \\
       &       & Clipart & Infograph & Painting & Real & Avg & CUB & Pets & Dogs & NABird & Avg \\
\midrule
MoE & 1.1 & 69.1 \std{0.1} & 36.1 \std{0.2} & 65.7 \std{0.1} & 78.0 \std{0.3} & - & - & - & - & - & - \\
\midrule
Simple Avg & 1.0 & - & - & - & - & - & - & - & - & - & - \\
RegMean\cite{jin2023dataless} & 1.0 & 16.6 \std{3.7} & 5.8 \std{1.6} & 10.1 \std{3.8} & 15.8 \std{5.8} & - & - & - & - & - & - \\
ZipIt\cite{stoica2024zipit} & 1.0 & 29.7 \std{3.5} & 12.5 \std{1.0} & 27.3 \std{3.1} & 37.9 \std{2.2} & - & 47.8 \std{7.9} & 70.3 \std{7.6} & 38.6 \std{9.2} & 38.7 \std{4.6} & - \\
Git Re-Basin\cite{ainsworth2022git} & 1.0 & - & - & - & - & - & 35.0 \std{5.9} & 51.1 \std{15.6} & 26.4 \std{12.5} & 32.1 \std{1.7} & - \\
\method & 1.0 & 41.5 \std{2.6} & 17.4 \std{1.3} & 40.4 \std{2.9} & 56.2 \std{2.1} & - & 63.1 \std{2.6} & 68.3 \std{17.6} & 46.6 \std{17.7} & 63.2 \std{1.6} & - \\
\midrule
ZipIt & 1.2 & 51.8 \std{2.1} & 21.5 \std{1.0} & 46.7 \std{2.5} & 60.7 \std{2.3} & - & 69.4 \std{2.4} & 85.8 \std{1.8} & 62.5 \std{4.8} & 61.0 \std{1.8} & - \\
Permutation-Weight & 1.2 & - & - & - & - & - & 65.4 \std{3.8} & 83.7 \std{2.7} & 62.1 \std{7.0} & 61.8 \std{0.8} & - \\
\method & 1.2 & 57.5 \std{1.6} & 26.0 \std{0.9} & 54.3 \std{2.0} & 69.4 \std{1.1} & - & 72.0 \std{3.2} & 87.0 \std{2.8} & 70.7 \std{5.8} & 69.8 \std{0.8} & - \\
\midrule
ZipIt & 1.55 & 60.7 \std{2.0} & 25.8 \std{0.8} & 54.8 \std{2.3} & 67.7 \std{1.5} & - & 73.2 \std{1.0} & 90.1 \std{0.7} & 72.7 \std{1.5} & 67.2 \std{1.5} & - \\
Permutation-Weight & 1.55 & - & - & - & - & - & 72.1 \std{2.9} & 88.3 \std{1.0} & 70.9 \std{4.4} & 69.3 \std{0.7} & - \\
\method & 1.55 & 60.8 \std{1.4} & 28.3 \std{0.6} & 57.3 \std{2.0} & 72.0 \std{0.7} & - & 74.2 \std{2.9} & 90.4 \std{1.4} & 74.8 \std{3.1} & 72.5 \std{1.0} & - \\
\midrule
ZipIt & 1.8 & 63.0 \std{1.8} & 28.0 \std{1.0} & 57.7 \std{2.5} & 70.0 \std{1.7} & - & 74.9 \std{1.1} & 91.5 \std{0.7} & 76.3 \std{1.4} & 71.3 \std{1.1} & - \\
Permutation-Weight & 1.8 & - & - & - & - & - & 74.1 \std{2.4} & 90.2 \std{0.9} & 76.0 \std{3.8} & 72.5 \std{0.4} & - \\
\method & 1.8 & 63.4 \std{1.4} & 30.1 \std{0.5} & 60.2 \std{1.8} & 74.2 \std{0.7} & - & 77.3 \std{0.7} & 90.6 \std{1.4} & 76.7 \std{0.0} & 74.3 \std{0.6} & - \\
\midrule
Ensemble & 2.0 & 63.6 \std{1.3} & 30.3 \std{1.0} & 61.0 \std{1.6} & 74.7 \std{0.6} & - & 79.8 \std {1.4} & 92.7 \std {0.0} & 75.6 \std {0.5} & 80.5 \std {0.8} & - \\
\bottomrule
\end{tabular}}

\end{table}}

\def\AblationsAnimalsOracle#1{
\begin{table}[]
    \begin{tabular}{llllllll}
    \toprule
    Budget & Algorithm & Data Source & CUB & OxfordPets & NABird & StanfordDogs \\ \midrule
    \multirow{4}{*}{1.0} & \multirow{2}{*}{Weight Matching} & Imagenet & 51.3 \std {9.9} & 66.0 \std {6.2} & 49.7 \std {0.0} & 35.7 \std {9.0} \\ \cline{3-7}
    & & Original & 61.0 \std {7.7} & 65.7 \std {20.1} & 65.3 \std {0.3} & 43.9 \std {17.8} \\ \cline{2-7}
    & \multirow{2}{*}{Activation Matching} & Imagenet & 57.8 \std {1.0} & 69.8 \std {5.7} & 53.6 \std {1.7} & 55.1 \std {0.0} \\ \cline{3-7}
    & & Original & 64.3 \std {0.0} & 68.1 \std {18.8} & 62.6 \std {0.7} & 45.2 \std {22.9} \\ \hline

    \multirow{4}{*}{1.2} & \multirow{2}{*}{Weight Matching} & Imagenet & 67.3 \std {0.4} & 81.0 \std {1.1} & 65.0 \std {0.4} & 63.2 \std {0.0} \\ \cline{3-7}
    & & Original & 73.2 \std {2.6} & 86.7 \std {3.7} & 70.5 \std {0.3} & 71.7 \std {5.2} \\ \cline{2-7}
    & \multirow{2}{*}{Activation Matching} & Imagenet & 67.3 \std {0.0} & 87.1 \std {0.0} & 64.6 \std {0.0} & 69.4 \std {0.0} \\ \cline{3-7}
    & & Original & 72.8 \std {0.0} & 86.0 \std {2.1} & 69.9 \std {0.0} & nan \std {nan} \\ \hline

    \multirow{4}{*}{1.55} & \multirow{2}{*}{Weight Matching} & Imagenet & 68.0 \std {0.5} & 87.0 \std {1.1} & 67.1 \std {0.0} & 71.5 \std {0.6} \\ \cline{3-7}
    & & Original & 74.9 \std {2.7} & 89.2 \std {2.0} & 72.4 \std {0.7} & 74.7 \std {3.5} \\ \cline{2-7}
    & \multirow{2}{*}{Activation Matching} & Imagenet & 68.9 \std {0.6} & 89.0 \std {0.1} & 66.5 \std {0.5} & 72.9 \std {0.7} \\ \cline{3-7}
    & & Original & 74.4 \std {2.6} & 91.3 \std {0.5} & 73.3 \std {0.0} & 77.0 \std {0.3} \\ \hline

    \multirow{4}{*}{1.8} & \multirow{2}{*}{Weight Matching} & Imagenet & 69.2 \std {0.4} & 87.5 \std {0.0} & 68.0 \std {0.3} & 74.0 \std {0.3} \\ \cline{3-7}
    & & Original & 75.3 \std {2.5} & 91.7 \std {0.0} & 74.1 \std {0.7} & 77.2 \std {2.6} \\ \cline{2-7}
    & \multirow{2}{*}{Activation Matching} & Imagenet & 69.5 \std {0.3} & 89.1 \std {0.6} & 67.5 \std {0.3} & 74.7 \std {0.2} \\ \cline{3-7}
    & & Original & 75.9 \std {2.1} & 90.3 \std {1.4} & 75.0 \std {0.0} & 79.5 \std {0.9} \\ \bottomrule
    \end{tabular}
    \caption{Data ablation for other datasets}
    \label{tab:animal_ablation}
\end{table}
}

\def\AblationsAnimals#1{
\begin{table}[]
    \begin{tabular}{llllllll}
    \toprule
    Budget & Algorithm & Data Source & CUB & OxfordPets & NABird & StanfordDogs \\ \midrule
    \multirow{4}{*}{1.0} & \multirow{2}{*}{Weight Matching} & Imagenet & nan \std{nan} & 80.1 \std{0.0} & 69.1 \std{0.0} & nan \std{nan} \\ \cline{3-7}
    & & Original & 73.4 \std{1.8} & 84.1 \std{2.9} & 70.6 \std{0.7} & 65.8 \std{3.4} \\ \cline{2-7}
    & \multirow{2}{*}{Activation Matching} & Imagenet & 69.3 \std{0.0} & nan \std{nan} & nan \std{nan} & 63.8 \std{0.0} \\ \cline{3-7}
    & & Original & 77.3 \std{2.5} & 84.3 \std{2.6} & 69.8 \std{1.0} & 61.8 \std{0.0} \\ \midrule

    \multirow{4}{*}{1.2} & \multirow{2}{*}{Weight Matching} & Imagenet & nan \std{nan} & 87.5 \std{0.0} & 70.1 \std{0.6} & 70.5 \std{0.0} \\ \cline{3-7}
    & & Original & 78.1 \std{2.1} & 90.6 \std{0.0} & 71.8 \std{0.7} & 74.4 \std{1.6} \\ \cline{2-7}
    & \multirow{2}{*}{Activation Matching} & Imagenet & nan \std{nan} & nan \std{nan} & nan \std{nan} & nan \std{nan} \\ \cline{3-7}
    & & Original & 77.2 \std{2.4} & 89.1 \std{0.8} & 71.6 \std{0.8} & 73.9 \std{1.9} \\ \midrule

    \multirow{4}{*}{1.55} & \multirow{2}{*}{Weight Matching} & Imagenet & 75.0 \std{0.0} & 90.8 \std{0.0} & 71.6 \std{0.5} & 75.7 \std{0.8} \\ \cline{3-7}
    & & Original & 78.6 \std{2.2} & 91.4 \std{0.5} & 73.5 \std{0.5} & 77.4 \std{0.9} \\ \cline{2-7}
    & \multirow{2}{*}{Activation Matching} & Imagenet & 74.8 \std{0.0} & nan \std{nan} & nan \std{nan} & 75.4 \std{0.0} \\ \cline{3-7}
    & & Original & 78.6 \std{2.2} & 91.5 \std{0.9} & 73.4 \std{0.5} & 77.4 \std{1.2} \\ \midrule

    \multirow{4}{*}{1.8} & \multirow{2}{*}{Weight Matching} & Imagenet & 75.9 \std{0.0} & 92.2 \std{0.0} & 73.0 \std{0.0} & 77.6 \std{0.0} \\ \cline{3-7}
    & & Original & 76.5 \std{0.0} & 91.9 \std{0.0} & 74.9 \std{0.0} & 79.9 \std{0.0} \\ \cline{2-7}
    & \multirow{2}{*}{Activation Matching} & Imagenet & 76.3 \std{0.0} & 92.3 \std{0.0} & 72.6 \std{0.0} & 77.3 \std{0.0} \\ \cline{3-7}
    & & Original & 76.8 \std{0.0} & nan \std{nan} & nan \std{nan} & 79.2 \std{0.0} \\ \midrule
    \end{tabular}
\end{table}

    \end{tabular}
    \caption{Data ablation for other datasets}
    \label{tab:animal_ablation}
\end{table}
}

\def\AblationsDomainNet#1{
\begin{table}[]
    \centering
    \begin{tabular}{llllllll}
    \toprule
    Budget & Algorithm & Data Source & Clipart & Infograph & Painting & Real \\ \midrule
    \multirow{4}{*}{1.0} & \multirow{2}{*}{Weight Matching} & Imagenet & 47.3 \std{1.2} & 20.5 \std{0.1} & 39.8 \std{0.0} & 55.1 \std{0.5} \\ \cline{3-7}
    & & Original & 45.9 \std{2.3} & 18.3 \std{1.8} & 42.7 \std{2.1} & 56.5 \std{0.5} \\ \cline{2-7}
    & \multirow{2}{*}{Activation Matching} & Imagenet & 44.4 \std{1.4} & 18.0 \std{1.3} & 40.6 \std{0.2} & 55.2 \std{0.0} \\ \cline{3-7}
    & & Original & 44.2 \std{1.7} & 17.0 \std{1.2} & 41.5 \std{3.2} & 53.9 \std{0.6} \\ \hline

    \multirow{4}{*}{1.2} & \multirow{2}{*}{Weight Matching} & Imagenet & 59.2 \std{1.1} & 25.5 \std{0.2} & 53.8 \std{0.7} & 67.9 \std{0.0} \\ \cline{3-7}
    & & Original & 58.7 \std{1.8} & 27.0 \std{0.9} & 55.5 \std{2.4} & 69.1 \std{0.7} \\ \cline{2-7}
    & \multirow{2}{*}{Activation Matching} & Imagenet & 58.7 \std{1.3} & 26.4 \std{1.1} & 55.5 \std{2.8} & 68.5 \std{0.5} \\ \cline{3-7}
    & & Original & 57.7 \std{1.7} & 26.3 \std{1.4} & 55.0 \std{2.3} & 68.6 \std{0.4} \\ \hline

    \multirow{4}{*}{1.55} & \multirow{2}{*}{Weight Matching} & Imagenet & 62.1 \std{1.7} & 28.2 \std{0.5} & 58.0 \std{2.2} & 71.5 \std{0.4} \\ \cline{3-7}
    & & Original & 61.3 \std{1.8} & 28.6 \std{0.8} & 58.6 \std{2.0} & 71.6 \std{0.4} \\ \cline{2-7}
    & \multirow{2}{*}{Activation Matching} & Imagenet & 61.4 \std{1.8} & 28.6 \std{0.0} & 61.5 \std{0.0} & 71.6 \std{0.2} \\ \cline{3-7}
    & & Original & 61.1 \std{1.5} & 28.7 \std{1.0} & 58.1 \std{2.3} & 71.7 \std{0.2} \\ \hline

    \multirow{4}{*}{1.8} & \multirow{2}{*}{Weight Matching} & Imagenet & 63.2 \std{1.1} & 30.1 \std{0.9} & 60.7 \std{2.0} & 72.6 \std{0.3} \\ \cline{3-7}
    & & Original & 62.1 \std{1.5} & 30.3 \std{0.1} & 60.7 \std{1.5} & 72.9 \std{0.2} \\ \cline{2-7}
    & \multirow{2}{*}{Activation Matching} & Imagenet & 61.5 \std{0.2} & 29.0 \std{0.7} & 60.0 \std{1.6} & 73.1 \std{0.0} \\ \cline{3-7}
    & & Original & 62.4 \std{1.6} & 29.5 \std{0.7} & 60.0 \std{2.1} & 72.8 \std{0.1} \\ \bottomrule
    \end{tabular}
    \caption{Data ablation for DomainNet}
    \label{tab:domainnet_ablation}
\end{table}
}

\def\appTableDomainNet#1{
\begin{table*}[h]
    \centering
        \caption{\textbf{Detailed Results on DomainNet} We report the results for ResNet-50 here}
    \label{tab:domainnet_detailed}
    \resizebox{\linewidth}{!}{\begin{tabular}{lllllllllllllll}
\toprule
Method & Budget & Data & \multicolumn{2}{c}{in-re} &\multicolumn{2}{c}{cl-pa} &\multicolumn{2}{c}{cl-in} &\multicolumn{2}{c}{pa-in} &\multicolumn{2}{c}{cl-re} &\multicolumn{2}{c}{pa-re} \\
\midrule
PLeaS-Act & 1.2 & Original & 26.3 \std{1.2} & 69.8 \std{0.5} & 57.3 \std{0.7} & 53.6 \std{0.1} & 55.4 \std{0.0} & 26.6 \std{0.0} & 52.7 \std{0.2} & 25.6 \std{0.1} & 59.1 \std{1.3} & 69.2 \std{1.2} & 56.8 \std{1.2} & 70.3 \std{0.9} \\
PLeaS-Act & 1.2 & Imagenet & 26.6 \std{0.9} & 69.6 \std{0.8} & 56.4 \std{0.9} & 53.9 \std{0.2} & 55.8 \std{0.7} & 26.3 \std{0.5} & 53.3 \std{0.2} & 25.6 \std{0.4} & 59.1 \std{1.3} & 69.3 \std{0.9} & 56.7 \std{1.0} & 70.2 \std{0.9} \\
PLeaS-Act & 1.0 & Original & 17.4 \std{1.3} & 55.0 \std{1.5} & 40.2 \std{2.0} & 39.9 \std{1.1} & 39.5 \std{1.5} & 17.8 \std{0.8} & 35.9 \std{0.0} & 17.3 \std{0.0} & 42.8 \std{2.0} & 54.9 \std{2.0} & 42.3 \std{1.7} & 56.1 \std{1.2} \\
PLeaS-Act & 1.0 & Imagenet & 17.0 \std{1.3} & 51.9 \std{1.3} & 39.7 \std{2.4} & 38.8 \std{0.8} & 37.6 \std{2.1} & 16.6 \std{0.7} & 37.1 \std{1.6} & 16.5 \std{0.7} & 42.2 \std{2.9} & 53.9 \std{1.5} & 40.9 \std{2.6} & 54.4 \std{0.9} \\
PLeaS-Act & 1.55 & Original & 29.0 \std{0.8} & 72.4 \std{0.3} & 60.6 \std{0.6} & 57.3 \std{0.0} & 59.7 \std{0.6} & 28.7 \std{0.3} & 56.4 \std{0.1} & 27.9 \std{0.4} & 62.5 \std{0.8} & 72.3 \std{0.9} & 60.6 \std{1.0} & 72.7 \std{0.7} \\
PLeaS-Act & 1.55 & Imagenet & 29.0 \std{0.8} & 72.5 \std{0.6} & 60.3 \std{0.7} & 57.8 \std{0.5} & 59.9 \std{0.6} & 29.0 \std{0.2} & 56.9 \std{0.2} & 28.1 \std{0.3} & 62.4 \std{0.9} & 72.6 \std{0.7} & 60.1 \std{0.9} & 72.9 \std{0.8} \\
PLeaS-Act & 1.8 & Original & 29.5 \std{0.8} & 73.7 \std{0.3} & 62.3 \std{0.5} & 59.0 \std{0.2} & 61.3 \std{0.4} & 29.8 \std{0.2} & 58.3 \std{0.2} & 29.1 \std{0.0} & 63.7 \std{0.6} & 73.8 \std{0.6} & 62.0 \std{1.0} & 73.9 \std{0.7} \\
PLeaS-Act & 1.8 & Imagenet & 30.1 \std{0.8} & 73.6 \std{0.6} & 62.1 \std{0.2} & 59.9 \std{0.4} & 61.6 \std{0.3} & 29.9 \std{0.1} & 58.4 \std{0.2} & 28.8 \std{0.1} & 63.7 \std{0.7} & 74.0 \std{0.7} & 61.7 \std{0.5} & 73.9 \std{0.7} \\
Permutation-Act & 1.2 & Original & 22.3 \std{0.9} & 62.9 \std{1.2} & 50.7 \std{1.2} & 45.9 \std{0.2} & 48.8 \std{0.0} & 22.1 \std{0.0} & 45.4 \std{0.4} & 21.9 \std{0.4} & 51.4 \std{1.0} & 61.6 \std{1.7} & 49.1 \std{1.3} & 63.9 \std{1.1} \\
Permutation-Act & 1.0 & Original & 7.6 \std{0.2} & 24.5 \std{1.9} & 15.4 \std{1.7} & 15.5 \std{1.8} & 15.6 \std{0.5} & 7.6 \std{0.5} & 15.4 \std{0.9} & 6.8 \std{0.0} & 17.1 \std{1.4} & 24.6 \std{1.7} & 18.7 \std{0.6} & 26.6 \std{2.0} \\
Permutation-Act & 1.55 & Original & 27.1 \std{0.7} & 69.7 \std{0.7} & 58.2 \std{1.0} & 53.6 \std{0.4} & 56.6 \std{0.6} & 26.8 \std{0.1} & 53.3 \std{0.4} & 25.9 \std{0.3} & 59.3 \std{1.1} & 68.8 \std{0.8} & 57.2 \std{0.8} & 70.3 \std{1.0} \\
Permutation-Act & 1.8 & Original & 28.9 \std{0.8} & 71.8 \std{0.6} & 60.7 \std{0.6} & 56.8 \std{0.3} & 59.7 \std{0.6} & 28.6 \std{0.3} & 56.5 \std{0.4} & 27.7 \std{0.5} & 62.4 \std{1.0} & 71.4 \std{0.8} & 60.2 \std{0.8} & 72.4 \std{0.9} \\
PLeaS-Weight & 1.2 & Original & 27.3 \std{1.0} & 70.5 \std{0.5} & 58.8 \std{0.2} & 54.6 \std{0.3} & 56.6 \std{0.8} & 27.2 \std{0.3} & 53.5 \std{0.1} & 26.1 \std{0.5} & 60.9 \std{0.2} & 68.8 \std{0.1} & 58.3 \std{0.9} & 70.5 \std{0.8} \\
PLeaS-Weight & 1.2 & Imagenet & 27.4 \std{0.9} & 70.2 \std{0.6} & 57.5 \std{0.9} & 54.3 \std{0.3} & 56.9 \std{0.4} & 27.3 \std{0.3} & 53.7 \std{0.4} & 26.0 \std{0.6} & 59.6 \std{1.0} & 69.9 \std{1.1} & 57.6 \std{0.8} & 70.8 \std{0.9} \\
PLeaS-Weight & 1.0 & Original & 19.2 \std{2.0} & 56.8 \std{1.4} & 45.6 \std{0.2} & 39.9 \std{0.2} & 41.0 \std{1.5} & 19.0 \std{1.5} & 39.6 \std{1.3} & 17.1 \std{0.9} & 45.7 \std{3.5} & 57.3 \std{2.1} & 43.2 \std{2.5} & 58.1 \std{1.4} \\
PLeaS-Weight & 1.0 & Imagenet & 17.8 \std{1.9} & 55.0 \std{1.2} & 41.0 \std{2.9} & 40.8 \std{1.0} & 40.2 \std{1.8} & 17.9 \std{1.2} & 38.7 \std{1.4} & 16.7 \std{0.7} & 43.7 \std{3.7} & 56.4 \std{1.7} & 40.7 \std{2.7} & 56.1 \std{0.7} \\
PLeaS-Weight & 1.55 & Original & 28.5 \std{0.9} & 72.9 \std{0.4} & 61.3 \std{0.0} & 58.2 \std{0.0} & 60.0 \std{0.4} & 29.1 \std{0.2} & 56.7 \std{0.1} & 28.1 \std{0.2} & 63.2 \std{0.6} & 72.1 \std{0.7} & 60.4 \std{0.8} & 73.0 \std{0.6} \\
PLeaS-Weight & 1.55 & Imagenet & 29.1 \std{1.1} & 72.7 \std{0.5} & 60.4 \std{1.0} & 57.4 \std{0.6} & 60.4 \std{0.2} & 29.4 \std{0.4} & 56.9 \std{0.3} & 28.1 \std{0.4} & 62.9 \std{0.7} & 72.2 \std{0.8} & 60.6 \std{0.9} & 72.9 \std{0.6} \\
PLeaS-Weight & 1.8 & Original & 29.7 \std{1.1} & 73.8 \std{0.4} & 62.7 \std{0.2} & 59.4 \std{0.3} & 61.2 \std{0.2} & 30.1 \std{0.2} & 58.2 \std{0.4} & 29.1 \std{0.3} & 64.1 \std{0.5} & 73.2 \std{0.5} & 62.3 \std{0.9} & 73.8 \std{0.7} \\
PLeaS-Weight & 1.8 & Imagenet & 29.7 \std{1.0} & 74.0 \std{0.5} & 62.0 \std{0.8} & 58.9 \std{0.5} & 61.5 \std{0.1} & 30.3 \std{0.1} & 58.7 \std{0.1} & 29.1 \std{0.1} & 63.6 \std{0.7} & 73.7 \std{0.5} & 61.7 \std{0.6} & 74.4 \std{0.6} \\
Permutation-Weight & 1.2 & Original & 24.8 \std{1.1} & 65.0 \std{0.7} & 55.0 \std{0.0} & 48.4 \std{0.0} & 51.8 \std{0.8} & 24.0 \std{0.3} & 48.5 \std{0.9} & 23.3 \std{0.3} & 56.3 \std{0.3} & 63.7 \std{0.2} & 52.3 \std{0.5} & 65.5 \std{1.1} \\
Permutation-Weight & 1.0 & Original & 10.5 \std{0.7} & 34.6 \std{2.7} & 28.3 \std{0.1} & 22.9 \std{0.1} & 21.8 \std{1.2} & 9.7 \std{0.7} & 21.1 \std{1.9} & 9.4 \std{0.1} & 29.1 \std{1.3} & 38.1 \std{2.9} & 26.5 \std{1.5} & 38.1 \std{1.2} \\
Permutation-Weight & 1.55 & Original & 27.5 \std{0.9} & 69.7 \std{0.7} & 59.9 \std{0.2} & 54.9 \std{0.1} & 57.7 \std{0.8} & 27.7 \std{0.5} & 54.1 \std{0.7} & 26.3 \std{0.2} & 60.6 \std{0.8} & 69.4 \std{0.8} & 57.4 \std{0.9} & 70.4 \std{0.9} \\
Permutation-Weight & 1.8 & Original & 28.9 \std{0.8} & 71.7 \std{0.5} & 61.6 \std{0.1} & 57.3 \std{0.1} & 60.5 \std{0.7} & 29.0 \std{0.3} & 56.8 \std{0.7} & 28.1 \std{0.6} & 63.1 \std{0.9} & 71.1 \std{0.7} & 60.8 \std{0.8} & 72.2 \std{1.0} \\
ZipIt! & 1.2 & Original & 60.0 \std{1.2} & 21.9 \std{1.1} & 52.4 \std{1.4} & 44.8 \std{0.7} & 49.8 \std{0.3} & 21.4 \std{0.6} & 45.7 \std{0.8} & 20.6 \std{0.2} & 54.1 \std{2.2} & 59.6 \std{1.3} & 50.3 \std{1.3} & 63.6 \std{0.7} \\
ZipIt! & 1.0 & Original & 35.8 \std{1.3} & 13.2 \std{0.6} & 29.3 \std{2.6} & 26.1 \std{0.5} & 26.1 \std{1.2} & 12.4 \std{0.8} & 24.4 \std{1.2} & 10.9 \std{0.7} & 33.4 \std{2.8} & 37.0 \std{0.5} & 31.0 \std{1.8} & 39.4 \std{1.5} \\
ZipIt! & 1.55 & Original & 66.3 \std{0.7} & 26.6 \std{1.0} & 61.2 \std{1.0} & 53.6 \std{0.2} & 58.6 \std{0.2} & 25.9 \std{0.3} & 52.6 \std{0.4} & 25.5 \std{0.3} & 62.9 \std{1.1} & 67.5 \std{0.3} & 58.3 \std{1.1} & 69.9 \std{0.2} \\
ZipIt! & 1.8 & Original & 68.4 \std{0.1} & 28.7 \std{1.2} & 63.1 \std{0.7} & 56.6 \std{0.4} & 60.8 \std{0.3} & 27.9 \std{0.3} & 55.1 \std{0.7} & 27.3 \std{0.2} & 65.0 \std{0.6} & 69.8 \std{0.2} & 61.0 \std{0.9} & 72.5 \std{0.6} \\
\bottomrule
\end{tabular}}
\end{table*}
}

\def\appTableAnimals#1{
\begin{table*}[h]
    \centering
        \caption{\textbf{Detailed Results on Different Label Spaces} We report the results for ResNet-50 here}
    \label{tab:animals_detailed}
    \resizebox{\linewidth}{!}{\begin{tabular}{lllllllllllllll}
\toprule
Method & Budget & Data & \multicolumn{2}{c}{na-ox} &\multicolumn{2}{c}{na-st} &\multicolumn{2}{c}{cu-na} &\multicolumn{2}{c}{cu-st} &\multicolumn{2}{c}{cu-ox} &\multicolumn{2}{c}{st-ox} \\
\midrule
PLeaS-Act & 1.2 & Imagenet & 69.2 \std{0.2} & 88.6 \std{0.2} & 67.0 \std{0.3} & 68.6 \std{0.5} & 78.5 \std{0.6} & 69.4 \std{0.3} & 71.9 \std{0.5} & 71.3 \std{0.7} & 75.0 \std{1.1} & 89.4 \std{0.7} & 77.9 \std{0.4} & 90.5 \std{0.4} \\
PLeaS-Act & 1.2 & Original & 71.6 \std{0.2} & 89.5 \std{0.2} & 70.0 \std{0.5} & 71.0 \std{1.0} & 80.1 \std{0.5} & 72.0 \std{0.3} & 74.0 \std{0.5} & 75.2 \std{0.6} & 76.4 \std{0.8} & 90.2 \std{0.5} & 79.9 \std{0.5} & 91.9 \std{0.6} \\
PLeaS-Act & 1.0 & Imagenet & 66.2 \std{0.7} & 80.3 \std{1.1} & 63.6 \std{0.8} & 56.1 \std{0.5} & 76.6 \std{0.4} & 65.4 \std{0.6} & 67.4 \std{1.2} & 62.5 \std{1.4} & 70.9 \std{0.6} & 84.1 \std{1.1} & 73.1 \std{0.8} & 87.4 \std{0.3} \\
PLeaS-Act & 1.0 & Original & 70.2 \std{0.2} & 81.1 \std{0.2} & 68.2 \std{0.5} & 61.2 \std{0.6} & 79.7 \std{0.4} & 69.8 \std{0.4} & 71.5 \std{0.7} & 69.3 \std{1.2} & 74.4 \std{0.6} & 87.2 \std{0.5} & 78.2 \std{0.4} & 90.6 \std{0.4} \\
PLeaS-Act & 1.8 & Imagenet & 73.4 \std{0.2} & 91.6 \std{0.2} & 71.3 \std{0.4} & 75.7 \std{0.2} & 80.2 \std{0.3} & 72.3 \std{0.4} & 75.9 \std{0.4} & 77.4 \std{0.3} & 77.1 \std{1.1} & 91.8 \std{0.5} & 81.3 \std{0.3} & 92.3 \std{0.4} \\
PLeaS-Act & 1.8 & Original & 75.3 \std{0.3} & 91.3 \std{0.2} & 74.0 \std{0.6} & 76.9 \std{0.3} & 81.5 \std{0.4} & 75.0 \std{0.3} & 77.0 \std{0.4} & 79.5 \std{0.4} & 78.8 \std{0.5} & 92.4 \std{0.5} & 82.9 \std{0.5} & 92.8 \std{0.4} \\
PLeaS-Act & 1.55 & Imagenet & 71.5 \std{0.3} & 90.7 \std{1.1} & 69.8 \std{0.7} & 73.7 \std{1.1} & 79.5 \std{0.5} & 71.1 \std{0.6} & 74.9 \std{0.3} & 75.4 \std{0.4} & 76.3 \std{0.8} & 91.2 \std{0.5} & 80.3 \std{0.5} & 91.7 \std{0.3} \\
PLeaS-Act & 1.55 & Original & 73.2 \std{0.2} & 90.4 \std{0.2} & 72.5 \std{0.4} & 75.5 \std{0.7} & 81.2 \std{0.6} & 73.8 \std{0.3} & 75.9 \std{0.4} & 78.3 \std{0.6} & 78.0 \std{0.3} & 92.1 \std{0.5} & 81.9 \std{0.4} & 92.8 \std{0.5} \\
RegMean & 1.0 & Original & 44.1 \std{0.2} & 56.0 \std{0.2} & 51.8 \std{14.8} & 39.0 \std{16.4} & 54.1 \std{10.1} & 42.5 \std{13.4} & 45.5 \std{21.2} & 37.7 \std{22.3} & 55.1 \std{20.8} & 61.7 \std{21.4} & 37.4 \std{15.5} & 56.4 \std{16.2} \\
PLeaS-Weight & 1.2 & Imagenet & 71.0 \std{0.2} & 88.1 \std{0.2} & 68.5 \std{1.0} & 70.1 \std{0.5} & 79.2 \std{0.6} & 70.4 \std{0.3} & 73.7 \std{1.2} & 73.4 \std{0.7} & 73.5 \std{1.8} & 89.1 \std{0.5} & 77.5 \std{0.6} & 89.9 \std{0.3} \\
PLeaS-Weight & 1.2 & Original & 72.5 \std{0.2} & 88.5 \std{0.2} & 70.9 \std{0.5} & 72.5 \std{0.5} & 80.5 \std{0.4} & 72.5 \std{0.3} & 75.2 \std{0.8} & 76.4 \std{0.4} & 77.2 \std{0.9} & 90.7 \std{0.6} & 79.9 \std{0.3} & 91.5 \std{0.9} \\
PLeaS-Weight & 1.0 & Imagenet & 68.6 \std{0.5} & 79.4 \std{1.0} & 66.2 \std{0.8} & 58.9 \std{0.6} & 76.5 \std{0.6} & 65.8 \std{1.0} & 69.1 \std{1.3} & 63.9 \std{0.7} & 71.7 \std{1.1} & 83.7 \std{0.7} & 70.9 \std{0.4} & 84.6 \std{0.7} \\
PLeaS-Weight & 1.0 & Original & 70.6 \std{0.3} & 79.2 \std{0.4} & 69.6 \std{0.6} & 62.2 \std{0.4} & 79.4 \std{0.6} & 69.8 \std{0.4} & 72.4 \std{0.7} & 69.1 \std{0.5} & 75.2 \std{0.8} & 87.0 \std{0.4} & 76.1 \std{0.4} & 89.4 \std{0.5} \\
PLeaS-Weight & 1.8 & Imagenet & 73.4 \std{0.6} & 91.7 \std{0.5} & 71.2 \std{0.7} & 75.8 \std{0.5} & 80.0 \std{0.6} & 72.2 \std{0.2} & 75.5 \std{0.8} & 77.2 \std{0.5} & 76.1 \std{0.7} & 92.0 \std{0.7} & 80.9 \std{0.5} & 91.9 \std{0.6} \\
PLeaS-Weight & 1.8 & Original & 74.8 \std{0.2} & 91.6 \std{0.2} & 73.9 \std{0.3} & 77.5 \std{0.6} & 81.7 \std{0.4} & 74.8 \std{0.4} & 76.5 \std{0.5} & 80.2 \std{0.3} & 78.8 \std{0.4} & 92.3 \std{0.5} & 82.8 \std{0.5} & 92.9 \std{0.5} \\
PLeaS-Weight & 1.55 & Imagenet & 72.5 \std{0.2} & 89.9 \std{0.3} & 70.1 \std{1.0} & 74.1 \std{0.8} & 79.4 \std{0.6} & 71.6 \std{0.3} & 74.9 \std{0.6} & 75.8 \std{0.7} & 75.8 \std{1.1} & 91.0 \std{0.7} & 79.8 \std{0.7} & 91.4 \std{0.5} \\
PLeaS-Weight & 1.55 & Original & 73.9 \std{0.4} & 90.0 \std{0.2} & 72.8 \std{0.5} & 76.0 \std{0.6} & 81.4 \std{0.3} & 74.0 \std{0.3} & 76.1 \std{0.5} & 78.7 \std{0.5} & 77.9 \std{0.7} & 91.9 \std{0.4} & 81.7 \std{0.5} & 92.3 \std{0.7} \\
SimpleAvg & 1.0 & Original & 5.2 \std{0.9} & 21.3 \std{1.7} & 5.0 \std{0.7} & 11.2 \std{0.9} & 8.7 \std{0.5} & 4.1 \std{0.4} & 6.8 \std{1.0} & 9.0 \std{0.4} & 6.9 \std{0.8} & 18.0 \std{1.2} & 8.7 \std{0.4} & 17.9 \std{0.6} \\
Permutation-Weight & 1.2 & Original & 67.9 \std{0.7} & 87.5 \std{0.7} & 66.8 \std{0.3} & 69.6 \std{0.4} & 78.2 \std{0.5} & 68.3 \std{0.3} & 72.0 \std{0.8} & 72.3 \std{0.5} & 73.9 \std{0.9} & 88.9 \std{0.3} & 77.2 \std{0.7} & 89.7 \std{0.7} \\
Permutation-Weight & 1.0 & Original & 60.6 \std{0.4} & 78.1 \std{0.5} & 57.0 \std{0.7} & 59.1 \std{0.5} & 72.3 \std{0.6} & 59.9 \std{0.3} & 59.8 \std{0.8} & 60.4 \std{0.8} & 66.5 \std{0.6} & 82.1 \std{0.8} & 67.0 \std{0.7} & 82.5 \std{0.5} \\
Permutation-Weight & 1.8 & Original & 74.0 \std{0.6} & 91.0 \std{0.7} & 73.0 \std{0.4} & 76.3 \std{0.5} & 81.2 \std{0.3} & 73.8 \std{0.2} & 76.0 \std{0.5} & 78.5 \std{0.5} & 77.7 \std{0.6} & 92.1 \std{0.6} & 82.2 \std{0.4} & 92.5 \std{1.0} \\
Permutation-Weight & 1.55 & Original & 71.4 \std{0.6} & 90.4 \std{0.5} & 71.1 \std{0.4} & 73.9 \std{0.5} & 80.1 \std{0.5} & 71.9 \std{0.3} & 75.0 \std{0.7} & 76.2 \std{0.4} & 76.5 \std{0.9} & 91.0 \std{0.5} & 80.4 \std{0.3} & 91.9 \std{0.5} \\
Permutation-Act & 1.2 & Original & 67.1 \std{0.6} & 85.7 \std{0.6} & 64.6 \std{0.7} & 65.3 \std{0.9} & 77.4 \std{0.4} & 67.0 \std{0.5} & 70.2 \std{0.9} & 69.4 \std{0.8} & 73.0 \std{0.7} & 87.7 \std{0.4} & 76.7 \std{0.5} & 89.8 \std{0.4} \\
Permutation-Act & 1.0 & Original & 61.2 \std{0.7} & 75.9 \std{1.1} & 55.8 \std{0.7} & 54.8 \std{0.9} & 73.9 \std{0.4} & 61.2 \std{0.7} & 60.9 \std{0.8} & 59.4 \std{1.3} & 66.6 \std{2.7} & 80.0 \std{2.2} & 71.6 \std{0.5} & 86.4 \std{0.6} \\
Permutation-Act & 1.8 & Original & 74.1 \std{0.4} & 91.2 \std{0.5} & 73.0 \std{0.5} & 76.1 \std{0.8} & 81.0 \std{0.3} & 73.9 \std{0.4} & 76.4 \std{0.4} & 78.6 \std{0.5} & 78.5 \std{0.4} & 92.1 \std{0.6} & 82.4 \std{0.4} & 93.0 \std{0.4} \\
Permutation-Act & 1.55 & Original & 71.6 \std{0.4} & 89.6 \std{0.5} & 70.4 \std{0.3} & 72.4 \std{0.9} & 79.7 \std{0.4} & 71.5 \std{0.2} & 74.4 \std{0.5} & 75.5 \std{0.5} & 76.3 \std{0.5} & 90.9 \std{0.4} & 80.9 \std{0.5} & 92.3 \std{0.6} \\
ZipIt! & 1.2 & Original & 62.6 \std{0.4} & 85.0 \std{0.6} & 62.0 \std{0.9} & 63.1 \std{0.8} & 74.9 \std{0.7} & 63.6 \std{0.7} & 70.3 \std{0.7} & 66.5 \std{0.4} & 71.1 \std{0.9} & 85.1 \std{0.6} & 72.2 \std{0.5} & 88.2 \std{0.3} \\
ZipIt! & 1.0 & Original & 56.7 \std{0.6} & 76.7 \std{1.3} & 55.1 \std{0.6} & 52.3 \std{1.1} & 72.9 \std{0.4} & 57.8 \std{0.4} & 64.6 \std{0.6} & 58.3 \std{0.9} & 67.1 \std{0.6} & 80.8 \std{0.3} & 67.6 \std{1.3} & 85.5 \std{0.9} \\
ZipIt! & 1.8 & Original & 70.0 \std{0.4} & 91.0 \std{0.4} & 69.4 \std{0.8} & 75.0 \std{0.7} & 78.9 \std{0.6} & 70.4 \std{0.5} & 75.7 \std{0.4} & 76.0 \std{0.5} & 75.9 \std{0.7} & 91.0 \std{0.5} & 78.7 \std{0.6} & 92.2 \std{0.6} \\
ZipIt! & 1.55 & Original & 67.0 \std{0.4} & 90.0 \std{0.7} & 66.4 \std{0.3} & 72.9 \std{0.5} & 77.7 \std{0.4} & 67.3 \std{0.8} & 73.7 \std{0.6} & 73.2 \std{0.5} & 74.0 \std{0.5} & 89.9 \std{0.6} & 76.6 \std{0.7} & 91.4 \std{0.5} \\
\bottomrule
\end{tabular}}
\end{table*}
}

\def\mainTableViT#1{

\begin{table}[h!]
\centering

\resizebox{\linewidth}{!}{%
\begin{tabular}{lllllll}
\toprule
\multirow{2.3}{*}{Method} & \multicolumn{2}{c}{CIFAR-50+50} & \multicolumn{4}{c}{Other Datasets} \\
\cmidrule(lr){2-3} \cmidrule(lr){4-7} 
& Joint & Avg & NABirds & CUB & Pets & Dogs \\
\midrule
Simple Avg         & 72.6 & 84.5 & 7.8 & 65.9 & 86.1 & 60.4 \\
RegMean         & 72.7 & 84.7 & 7.7 & 66.2 & 85.1 & 58.5 \\
MuDSC           & 72.8 & 84.9 & 8.0 & 66.1 & 86.1 & 60.6 \\
\method         & 73.3 & 85.1 & 8.3 & 66.7 & 86.5 & 61.6 \\
\bottomrule
\end{tabular}
}
\caption{\textbf{Merging ViT models:} We merge pairs of ViT models fine-tuned from the same initialization and report the performance.  \method{} can outperform baselines across datasets.}
\label{tab:vit_performance}
\end{table}
}
\label{sec:experiments}
We show the effectiveness of our method in merging models fine-tuned on different datasets with different initializations. We investigate the following research questions:
\begin{enumerate}
    
    \item How does \method{} compare with prior work in merging models to produce a model of the same size (\cref{sec:same_size})?
    \item What is the trade-off between size of the merged model and its performance for \method{} (\cref{sec:diff_size})?
    \item How does \method{} perform if one does not have access to the training data of the models being merged (\cref{sec:no_data})?
    \item How does \method{} perform while merging models fine-tuned from the same initial model (\cref{sec:same_initialization}), or different models trained on the same data (\cref{app:imagenet_results})?
    \item What is the impact of varying the objective in \cref{eqn:perm_reg} on the performance of \method{}(\cref{app:ablations})?
\end{enumerate}

\subsection{Experimental Setup}\label{sec:setting}
\mainTableAverageSingleBudget{}

To obtain models for merging, we fine-tune ImageNet pre-trained ResNet models on other smaller datasets. We merge models trained (from \textit{different} initializations) on different data domains in a pair-wise fashion, and compute the accuracy of the merged model on both the data domains. For each domain, we average the accuracy across all such pairs.  
\subsubsection{Datasets}
Since we are dealing with classification models, we consider two sets of datasets (with shared and different label spaces) for training and merging models.

\textbf{Datasets with a shared label space.}
We fine-tune ImageNet pre-trained ResNet-50 models  on four different domains of the DomainNet~\cite{peng2019moment} dataset: Clipart, Infograph, Painting and Real. These datasets share a label space with 345 classes and comprise of images in various styles.

\textbf{Datasets with different label spaces.}
We fine-tune models on CUB~\cite{WahCUB_200_2011}, NABirds~\cite{van2015building}, Oxford-IIIT Pets~\cite{parkhi12a} and Stanford Dogs~\cite{KhoslaYaoJayadevaprakashFeiFei_FGVC2011} datasets, and merge them up to the penultimate layer.
Since the label spaces of these datasets are different, we aim to evaluate the representations of the penultimate layer of these merged models by training a linear probe on top of the representations. We average the results in the same manner as for DomainNet, and report the performance of different methods in \cref{tab:main_results}. In \cref{app:zipit_setting}, we follow the setting of \cite{stoica2024zipit}, using task specific heads from the original models to compute the accuracy of the merged model, which requires knowing the domain of each test data point. 

\subsubsection{Baselines}

We compare our method against prior works including Git Re-Basin~\cite{ainsworth2022git}, Simple Averaging~\cite{ilharco2023editing}, RegMean~\cite{jin2023dataless}, ZipIt!~\cite{stoica2024zipit} and MuDSC~\cite{xu2024training}. Note that RegMean has similar data and compute requirements as \methodnospace, and ZipIt! also supports partial merging of models like \methodnospace. We also consider two practical upper bounds --- training a router model based Mixture of Experts model (MoE), and ensembling the predictions (or activations) of the original models. The former requires storing both models and running one of them at inference in addition to the overhead of the router, hence having $1.1\times$ FLOPs and $2.1\times$ memory requirements, while the latter requires running both the models in parallel, and hence has $2\times$ FLOPs and memory requirements. We find that the performance of the ensemble and MoE models is close to the best performance of a single model on its training dataset.

For each task, we also report results for Permutations, which is the model obtained by weight averaging the component models after applying the permutations obtained from the first step of \methodnospace. Following the recommendation of REPAIR~\cite{jordan2023repair}, we recompute the batch-norm parameters of the model after merging for all methods. We run each merging experiment for three different seeds, and across two different initial models. We find that inter-run variation in performance is low, with the standard deviation usually being less than 1\%. We report disaggregated results along with these standard deviations in \cref{app:detailed_results}.

\subsection{Merging for the same size}
\label{sec:same_size}
\begin{figure*}[t]
    \centering
    \includegraphics[width=0.4\linewidth]{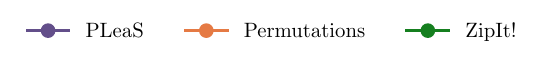}  %
    
    \begin{subfigure}[b]{0.32\textwidth}
        \centering
        \includegraphics[width=\linewidth,trim=0 1ex 0 1ex,clip]{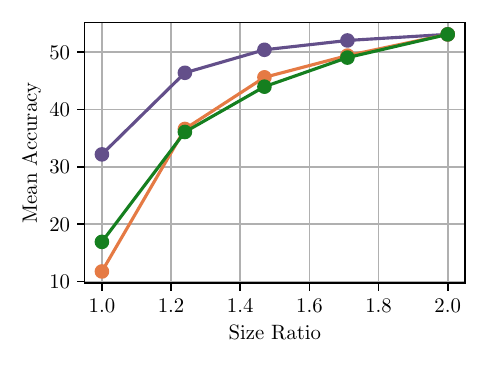}
        \caption{ResNet-18 (Shared)}
        \label{fig:same_rn18}
    \end{subfigure}
    \hfill
    \begin{subfigure}[b]{0.32\textwidth}
        \centering
        \includegraphics[width=\linewidth,trim=0 1ex 0 1ex,clip]{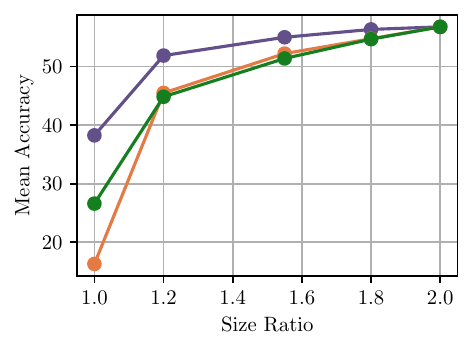}
        \caption{ResNet-50 (Shared)}
        \label{fig:same_rn50}
    \end{subfigure}
    \hfill
    \begin{subfigure}[b]{0.32\textwidth}
        \centering
        \includegraphics[width=\linewidth,trim=0 1ex 0 1ex,clip]{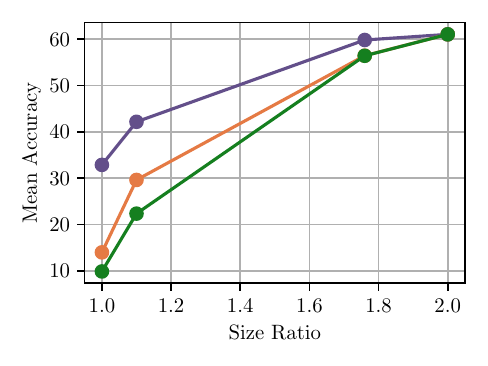}
        \caption{ResNet-101 (Shared)}
        \label{fig:same_rn101}
    \end{subfigure}

    \vspace{1em} %

    \begin{subfigure}[b]{0.32\textwidth}
        \centering
        \includegraphics[width=\linewidth,trim=0 1ex 0 1ex,clip]{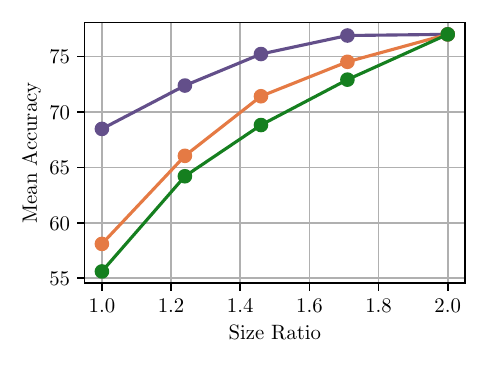}
        \caption{ResNet-18 (Different)}
        \label{fig:diff_rn18}
    \end{subfigure}
    \hfill
    \begin{subfigure}[b]{0.32\textwidth}
        \centering
        \includegraphics[width=\linewidth,trim=0 1ex 0 1ex,clip]{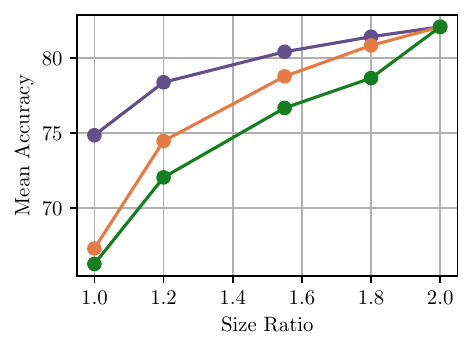}
        \caption{ResNet-50 (Different)}
        \label{fig:diff_rn50}
    \end{subfigure}
    \hfill
    \begin{subfigure}[b]{0.32\textwidth}
        \centering
        \includegraphics[width=\linewidth,trim=0 1ex 0 1ex,clip]{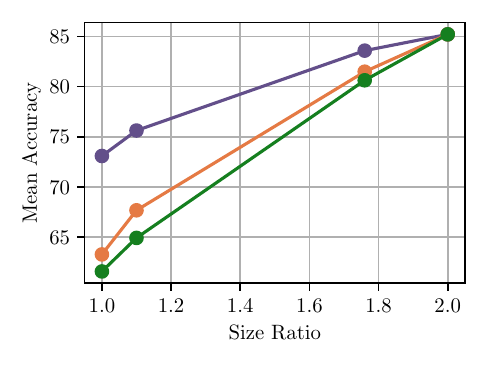}
        \caption{ResNet-101 (Different)}
        \label{fig:diff_rn101}
    \end{subfigure}

    \caption{\textbf{Memory-Performance trade-off for merged models}: We merge pairs of models fine-tuned on different datasets, and compute the average performance across all four datasets for two settings: datasets with a shared label space (top) and datasets with different label spaces (bottom). Plotting average accuracy against the final merged model size, we find that \method dominates the state-of-the-art methods.}\label{fig:mem_performance_tradeoff}
\end{figure*}

In \cref{tab:main_results}, we report the domain-wise performance for merged models at $1\times$ size of the original model for different methods. On DomainNet, we observe that \method{} outperforms the previous state-of-the-art method, MuDSC by over 8\% on average. Similarly, On datasets with different labels spaces, \method{} is better than MuDSC by almost 6.5\% on average.
Further, \method{} vastly outperforms RegMean, a method which which performs Least Squares without permutations and has a similar computation overhead as \methodnospace. This indicates that permuting the features before performing Least Squares is a crucial step which helps \method{} produce better models. Finally, These results also show the utility of the second step of \methodnospace, as it boosts the performance over Git Re-Basin by a large amount, nearly doubling the accuracy on DomainNet.

\subsection{Exploring the model size-accuracy tradeoff}\label{sec:size-accuracy-tradeoff}
\label{sec:diff_size}

We seek to demonstrate the effect of partial merging on the performance of the merged models. To do this, we merge models pairwise as above, and report the average of the performance of these merged models on their respective component domains for different sized ResNet models in \cref{fig:mem_performance_tradeoff}.

 We find that \method consistently outperforms ZipIt!\ at various compute/memory budgets and for all model scales. The gains are particularly striking for lower memory budgets, where \method outperforms ZipIt!\ by up to 10\% for ResNet-50 (~\cref{fig:same_rn50}). 
 The power of partial merging is also observed from these results, as one can see that increasing the flops by just 20\% leads to massive improvements in the accuracies in the shared label settings. 
These results also provide evidence of the effectiveness of our partial permutation scheme --- permutations can outperform ZipIt!\ at intermediate model budgets by up to 6\% (\eg for ResNet-101 with shared label spaces in~\cref{fig:same_rn101}).  We posit that this is because we can assign a non-uniform width multiplier across the layers of the merged model, which is important for larger models and those which are trained on disparate domains.  
As expected, the performance gap closes as the relative size of the merged models increases.

\subsection{Does \method need data from the training domains?}
\label{sec:no_data}
\begin{figure}[t]
    \centering
        \includegraphics[width=0.9\linewidth]{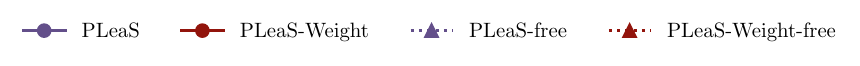}  %

\begin{subfigure}{0.49\linewidth}
        \centering
        \includegraphics[width=\linewidth]{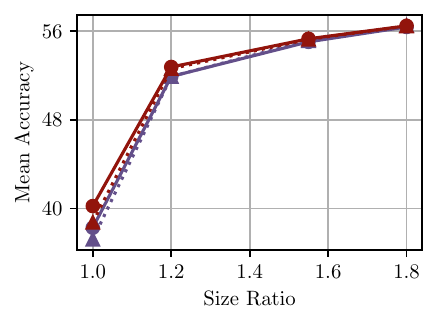}
        \subcaption{Shared Label Space}       
        \label{fig:data_same_label_spaces}
    \end{subfigure}
\begin{subfigure}{0.49\linewidth}
        \centering
        \includegraphics[width=\linewidth]{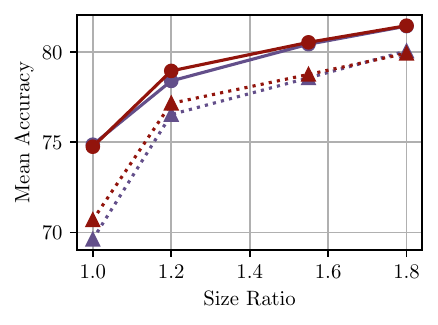}
        \subcaption{Different Label Spaces}
        \label{fig:data_diff_label_spaces}
    \end{subfigure}    
    \caption{\textbf{Investigating the data requirement of \method }: We run \method and \methoda using data from the actual domains or ImageNet (indicated by the suffix \texttt{free}) for both the Shared label space (\cref{fig:data_same_label_spaces}) and Different label spaces (\cref{fig:data_diff_label_spaces}) settings for ResNet-50. We plot the average accuracy across all datasets against the relative size of the output model. We find minimal performance drops for \methodw. }
     
    \label{fig:main_figure4}
\end{figure}

To investigate the data requirements of our method, we compare the performance of \method and \methodw when merging ResNet-50 models. We also compare the effect of using weight matching to find the permutations for \methodnospace, and we term this variant as \methodanospace. Note that this variant uses data only for the Least Squares step.
The performance-model size tradeoff is reported in \cref{fig:data_same_label_spaces,fig:data_diff_label_spaces} for shared and different label spaces.

We find that \methodw retains a similar performance when using ImageNet instead of the actual domain data for merging models on DomainNet, achieving a drop of less than 1\% in accuracy at $1\times$ model size. There is almost no drop at higher sizes of the merged model. 
Notably, even on the more difficult task of merging models with different label spaces, using ImageNet data for computing activations can perform competitively to using the actual data: linear probing on the representations from \methodw performs within 2\% of the \method at $1.2\times$ model size, and the gap is less than 4\% at $1\times$ model size. This result is particularly encouraging, since it extends the practical applicability of \methodw to scenarios where data from the training domains may not be available. Note that while we must use data from the actual domains for linear probing, i.e. to assess the quality of the representations, we do not use it for actually merging the models.
We also find that \methoda performs similarly to \method{} in the both the shared and different label space settings for ResNet-50.
Further, \methoda is less affected when ImageNet data is used, since the permutations computed by \methoda do not depend on the data.

\subsection{Merging models with the same initialization} 
\label{sec:same_initialization}

\sameInitTable{}
In Tab~\ref{tab:same_init}, we evaluate the performance of our method for merging ResNet-50 models fine-tuned from the same starting model. We compare against simple average (Task Vectors), MuDSC, and RegMean and find that the performance is similar across methods, with \method being slightly better than the baselines. In fact, task vectors is an effective baseline here. However, we note that 20\% extra parameters in the merged model can lead to closing the gap between the ensemble and the merged model produced by \methodnospace, demonstrating the need for flexible merging methods. 

\subsubsection{Merging ViTs}
\label{sec:architectures}

\mainTableViT{}
In \cref{tab:vit_performance}, we present results of merging CLIP style ViT models. In accordance with prior work~\cite{xu2024training}, we merge models starting from the same initialization. We consider two settings -- the different label space setting described in \cref{sec:setting} and CIFAR50+50 from \cite{stoica2024zipit, xu2024training}. In the latter, the 100 labels from CIFAR-100 are partitioned into 2 sets of 50 labels each, and a ViT is trained on each of these sets (using a CLIP-like loss). More details on the setup are described in \cref{app:exp_details}. The accuracy of the merged model is reported on both these partitions separately by considering only 50 classes at a time (denoted by Avg in the table), as well as on the Joint CIFAR-100 dataset (by considering all 100 classes together). Note that since we use CLIP-like models, we can use the language head directly for classification despite different label spaces. For the other datasets, we follow the protocol from \cref{sec:setting}.

We observe that \method{} boosts the performance slightly over the baselines. For example, it increases performance by 0.7\% on CIFAR-100 over Task Vectors. \method{} also out-performs MuDSC by around 0.6\% on the larger datasets. While these gains are small, they are non-trivial and are more than the boost reported by the previous state-of-the-art method (MuDSC). We believe this presents an exciting opportunity to further study the symmetries and invariances in transformers in order to merge them better.

\section{Limitations}
The scope of this study is limited to merging models with the \textit{same} architecture, and applying \method to merge different architectures could be an interesting future direction. Since \method is a two-stage algorithm, its running time is greater than some existing works~\cite{stoica2024zipit, ilharco2023editing, jin2023dataless}. However, since the second step can be computed in parallel for all layers, as discussed in \cref{app:costs}.

\section{Conclusion}
In this work, we present \methodnospace, an algorithm to merge models trained on different datasets starting from different initializations. We demonstrate that \method can effectively produce merged models at different points on the compute-performance trade-off curve. We also propose \methodwnospace, a variant which can merge models without needing any data from the training domains of the component models, and empirically validate that its performance is comparable to running \method with data, which widens its applicability to data-scarce regimes.  

\section*{Acknowledgement} 
This work is supported by Microsoft Grant for Customer Experience Innovation and the National Science Foundation under grant no.~2019844, 2112471, and 2229876. JH is supported by the NSF Graduate Research Fellowship Program. PWK is supported by the Singapore National Research Foundation and the National AI Group in the Singapore Ministry of Digital Development and Innovation under the AI Visiting Professorship Programme (award number AIVP-2024-001).

\small
\bibliographystyle{ieeenat_fullname}
\bibliography{references}
\clearpage

\appendix
\section{Experimental and Implementation details}
\label{app:exp_details}
In this section, we provide more details about our experiments. We conduct all experiments using PyTorch \cite{paszke2019pytorch}. We use two ImageNet pretrained base models for fine-tuning. One of these is the default \texttt{ResNet50\_Weights.IMAGENET1K\_V1} from PyTorch, while we pre-train the other starting from random initialization following the same pipeline. For fine-tuning the models on each domain, we use the Adam \cite{kingma2017adam} optimizer, and sweep the learning rates logarithmically between [1e-4,1e-1], testing out 4 values for LR. We validate on the validation subset wherever available, and on 10\% of the training dataset where an explicit val set is not provided. We use standard image augmentation techniques. Our MoE model has a light-weight router, which is a 3 layer CNN trained to predict which model to use for classifying an image.

For finding permutation symmetries, we use the official implementation of Git Re-Basin at \href{https://github.com/samuela/git-re-basin}{this url}. We also rely on the implementation of ZipIt!\ and MuDSC for the comparisons in \cref{sec:experiments}. 

For solving the least squares objective for \methodnospace, we use SGD with a batch size of 32, a learning rate of $10^{-3}$. We sample equally from both datasets in each batch for experiments involving data. We run our algorithm for 100 steps, and find that it converges quickly. For  the first step of \methodnospace, we similarly compute the activations on 100 batches of data for matching and finding the optimal permutations. We also reset batch norm parameters using 100 batches of data from the actual domains for all methods.

For evaluations concerning the same label space setting, we ensure that the final model produces a distribution over the output classes. For ZipIt!, we achieve this by ensembling the predictions across multiple task specific heads. \method on the other hand already produces models with the same output dimensions as the original models.

For evaluations on different label spaces, we train a linear probe on the final layer representations for each merged model. We use training data from the target domains to train this linear probe, run Adam with a learning rate of $10^{-3}$, with a batch size of 64 for 50 epochs.

As an example evaluation for Domainnet, we have 8 models, two each on Clipart, Infograph, Painting and Real domains. We merge these pair-wise. We hence have 12 (6 domain pairs and two models per pair) merged models. For each merged model, we compute the accuracy on its component domains. Hence, for each domain, we have 6 performances (3 domain pairs and two models per pair). We report the average of these 6 numbers in \cref{tab:main_results}.

\paragraph{ViT} To obtain models for merging with CIFAR-50+50 for \cref{sec:architectures}, we follow the protocol from \cite{xu2024training}. In particular, we use ImageNet pretrained ViT models, and train two such models on disjoint 50 class subsets of CIFAR-100. These are trained with CLIP language embeddings for the final layer. This process is repeated thrice to get different sets of classes and pairs of models. After merging, the performance of the merged model is measured on CIFAR-100 as well as the two subsets of classes that the moders were trained on. The average of these results is reported in \cref{tab:vit_performance}. For the other datasets, we start off with ViT-B/16 model from the OpenCLIP project\cite{ilharco_gabriel_2021_5143773} which is pretrained on the LAION-400M dataset. We fix the text encoder, and fine-tune the image encoder on various datasets. This fine-tuned model achieves an accuracy of (72\%, 91\%, 68\%, 24\%) on CUB, Pets, Dogs and NABirds datasets resp. We then merge the model pairwise and report the results.    

\subsection{Compute time and cost}
\label{app:costs}
All our experiments (apart from the pretraining and fine-tuning runs to get the original models) are run on a single RTX 2080 Ti GPU. The first step of our method runs in 2 minutes, with the majority of time devoted to computing the activations. This is commensurate with ZipIt!~\cite{stoica2024zipit} and Git Re-Basin \cite{ainsworth2022git} The second step takes around 4 minutes, which is similar to RegMean \cite{jin2023dataless}. We believe that this can be significantly reduced with better dataloading strategies and more efficient implementation, but that is beyond the scope of this paper.  
\subsection{Computing the permutation matrices}
We use the algorithms of Git Re-Basin to compute the permutation matrices \(P_i\).
For activation matching, we collect the representations for both models over a batch of data, and measure the alignment between two neurons as the squared distance between their representations (closer $\rightarrow$ better alignment). 
Then for each layer, we find the bipartite matching (i.e. permutation) between the two models that minimizes the total distance using a standard algorithm (\texttt{scipy.optimize.linear\_sum\_assignment}).
For weight matching, the alignment between two neurons is the squared distance between its input and output vectors.
This time, the permutations at adjacent layers interact, so we perform an alternating minimization, solving for the permutations one layer at a time until we reach a fixed point.
\subsection{Computing the layerwise merging ratio}
\label{app:qp}
Note that $k_i$ can be different for each layer. Given a configuration $K=\{\frac{k_i}{d_i}:i \in [L]\}$, we can model the FLOPs/memory of the merged model as a quadratic function of $k_i$, which we denote as $\text{Footprint}(K)$. For a given relative memory/FLOPs budget $B$, we want to find $K$ s.t. $\text{Footprint}(K) \leq B$ to maximize the accuracy of a model merged with the configuration $K$. We scale everything so that $B=1$ corresponds to the footprint of a single model. This problem is NP-Hard. We propose a relaxation of the problem in order to get an approximate solution. First, we measure the performance of a set of models merged with ``leave one out" configurations of $K$, where for each layer $i$, we construct $K_{i}^0 = \{k_j : k_j = d_i \ \text{if} \ j=i, \ 0 \ \text{otherwise}\}$ and $K_{i}^1 = \{k_j : k_j = 0 \ \text{if} \ j=i, \ d_i \ \text{otherwise}\}$. $K_i^0$ corresponds to merging only layer $i$, keeping all other layers unmerged, and $K_i^1$ corresponds to merging every other layer while keeping $i$ unmerged.  We also compute the accuracies of the fully merged model (denoted by $K^0$) and the ensemble (denoted by $K^1$). Then, we approximate the accuracy of any given $K$ with a linear function as 
\begin{align*}
\text{Acc}(K) = \sum_{i=1}^L \frac{k_i}{d_i} \bigg((2-B)(\text{Acc}(K^1)-\text{Acc}(K_{i}^0)) \\
- (1-B)(\text{Acc}(K_{i}^0) - \text{Acc}(K^0))\bigg)    
\end{align*}
This approximates the effect of $k_i$ on model performance at budget $B$ by linearly interpolating between the performance with fully merging layer $i$ and keeping it separate. We then propose to solve a quadratically constrained linear program to maximize $\text{Acc}(K)$ subject to $\text{Footprint}(K) \leq B$.
This program is non-convex however Gurobi \citep{gurobi} is able to solve the program to global optimality in a few seconds.
To faithfully compute the performance of the merged model, one would require validation samples from the target domain. However, we empirically observe that using the accuracy of a configuration $K$ on ImageNet is a good proxy for its performance on other merging tasks as well, and we hence use it to compute the layer-wise merging ratio for all our experiments. 
\subsubsection{Empirical Results}
\begin{figure}
  \begin{subfigure}{0.49\linewidth}
    \centering
    \includegraphics[width=\linewidth]{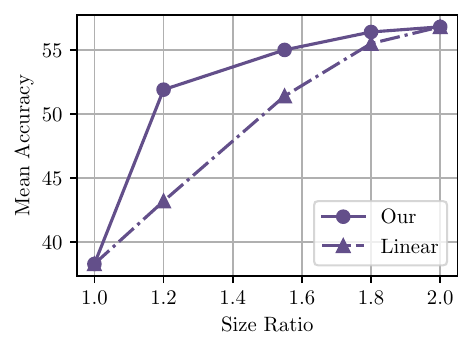}
    \subcaption{\method}
    \label{fig:scheduling_1}
  \end{subfigure}
  \hfill
  \begin{subfigure}{0.49\linewidth}
    \centering
    \includegraphics[width=\linewidth]{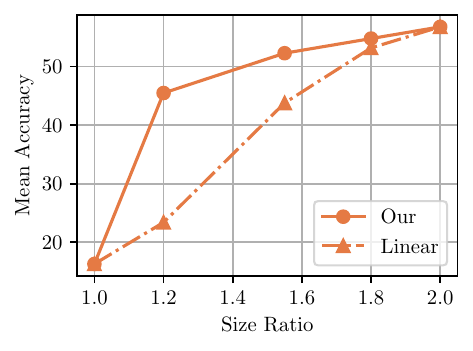}
    \subcaption{Permutations}
    \label{fig:scheduling_2}
  \end{subfigure}
  \caption{\textbf{Comparing our strategy for layer-wise merging with a linear baseline}: We merge models using \method{} and permutations using the strategy described in \cref{app:qp} and a linear strategy where $\frac{k}{d}$ is held constant. }
\end{figure}
In~\cref{fig:scheduling_1,fig:scheduling_2}, we compare the QP method with a baseline strategy which assigns the number of units in each layer to be a constant ratio. We find that our strategy outperforms this baseline for both \method{} and Permutations. 

\section{Additional Results}
\subsection{What to optimize for Least Squares?} 
\label{app:ablations}
\methodAblations{}
In \cref{eqn:perm_reg}, we propose to solve a least squares problem involving the permuted average activations from each layer of the component models. In \cref{tab:method_ablation}, we demonstrate that this choice is not only natural, but also performs better than other alternatives. It is also interesting to note that the second row in the table corresponds to a permuted version of RegMean\cite{jin2023dataless}. This formulation performs better than RegMean, indicating that using permutations is necessary to align features for networks which were differently initialized. Further, row 3 is similar to the objective proposed by \cite{NEURIPS2018_ad8e88c0}, but we show that \method outperforms this objective as well.

\subsection{Reducing the  accuracy barrier on ImageNet}
\label{app:imagenet_results}
\begin{figure}
  \begin{subfigure}{0.49\linewidth}
    \centering
    \includegraphics[width=\linewidth]{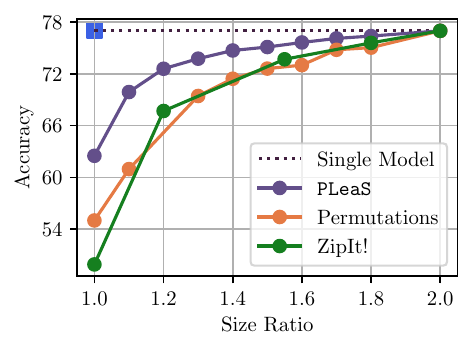}
    \subcaption{ImageNet performance}
    \label{fig:imagenet_consolidated}
  \end{subfigure}
  \hfill
  \begin{subfigure}{0.49\linewidth}
    \centering
    \includegraphics[width=\linewidth]{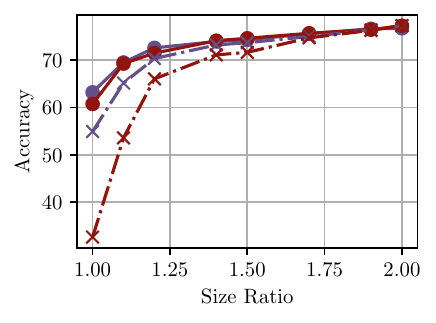}
    \subcaption{Data-free setting}
    \label{fig:imagenet_data}
  \end{subfigure}
  \caption{\textbf{Merging models trained on ImageNet}: In \cref{fig:imagenet_consolidated}, we demonstrate how \methodw can reduce the accuracy barrier by 8\% for merging independently trained ImageNet models. This is further reduced for larger target model sizes. In \cref{fig:imagenet_data}, we show the effect of using synthetic data for computing the activations on this task, and find that synthetic data is a viable alternative at larger model sizes.}
\end{figure}
In this section, we show the performance of \method while merging ResNet-50 models trained independently on ImageNet. The accuracy of a single model on this task is 77.5\%. As seen from \cref{fig:imagenet_consolidated}, current methods including ZipIt!~\cite{stoica2024zipit} and Git Re-Basin~\cite{ainsworth2022git} struggle on merging models for this task, with the accuracy of the merged model being significantly lower than the accuracy of a single model. This has been referred to as the accuracy barrier on ImageNet in prior work. \method makes some progress towards lowering this barrier, and improves over Git Re-Basin by over 9\% at $1.0\times$ FLOPs budget. 
For context, this accuracy is at par with that obtained by merging WideResNet-50 models with a width multiplier of 2 using Git Re-Basin. 
More promisingly, the flexibility afforded by partially permuting and merging models gives another avenue to lower the accuracy barrier, with a model of size $1.4\times$ having an accuracy barrier of 2\% with \methodnospace. However, further work is needed to reduce this accuracy barrier.
In \cref{fig:imagenet_data}, we compare using synthetic data from \cite{baradad2022procedural} for all purposes of activation computation while merging ImageNet trained models. We find that using \methodw with synthetic data can come close to using actual data, being within 1\% in terms of accuracy at $1.2\times$ model size. \\

\subsection{Detailed Results}
\label{app:detailed_results}
Each of our evaluation was run across three random restarts. These random restarts shuffle the data used for computing activations and merging the models. They also affect the initialization of the merged model. Each pair evaluation was also run twice, swapping the order of pre-trained models used for either of the datasets of the pair. We hence have 6 runs for each dataset pair. In \cref{tab:domainnet_detailed,tab:animals_detailed}, we provide the results for each dataset pair, reporting the average and standard deviation across the 6 runs.
\appTableDomainNet{}
\appTableAnimals{}
\subsection{Using task specific heads}
\label{app:zipit_setting}
\oracleTable{}
In \cref{tab:oracle_results}, we report the results computed using the protocol mentioned in \cite{stoica2024zipit}. We find that \method outperforms ZipIt!\ in this evaluation across model budgets.

\subsection{\method{} with weight and activation matching}
\begin{figure*}[ht!]
    \centering
    \includegraphics[width=0.4\linewidth]{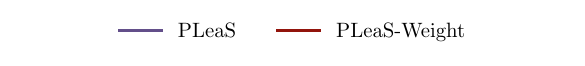}  %
    
    \begin{subfigure}[b]{0.32\textwidth}
        \centering
        \includegraphics[width=\linewidth,trim=0 1ex 0 1ex,clip]{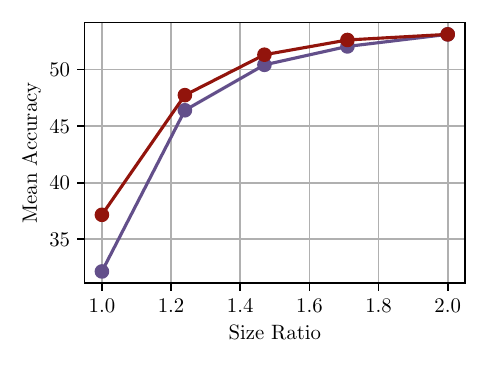}
        \caption{ResNet-18 (Shared)}
        \label{fig:same_wa_rn18}
    \end{subfigure}
    \hfill
    \begin{subfigure}[b]{0.32\textwidth}
        \centering
        \includegraphics[width=\linewidth,trim=0 1ex 0 1ex,clip]{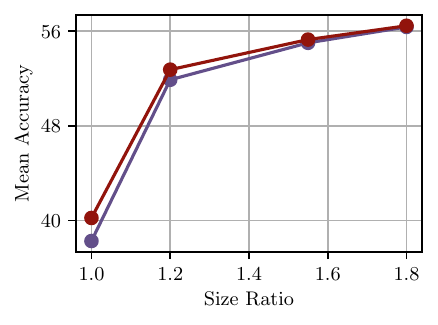}
        \caption{ResNet-50 (Shared)}
        \label{fig:same_wa_rn50}
    \end{subfigure}
    \hfill
    \begin{subfigure}[b]{0.32\textwidth}
        \centering
        \includegraphics[width=\linewidth,trim=0 1ex 0 1ex,clip]{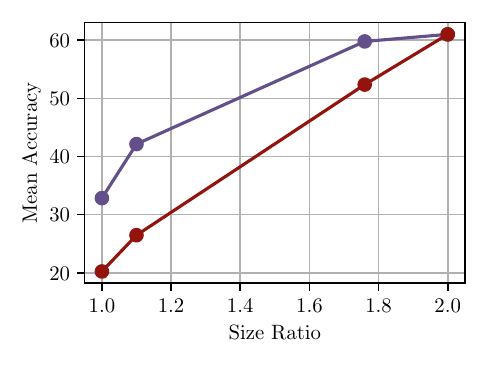}
        \caption{ResNet-101 (Shared)}
        \label{fig:same_wa_rn101}
    \end{subfigure}

    \vspace{1em} %

    \begin{subfigure}[b]{0.32\textwidth}
        \centering
        \includegraphics[width=\linewidth,trim=0 1ex 0 1ex,clip]{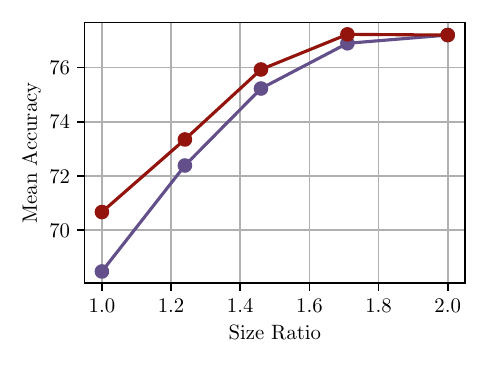}
        \caption{ResNet-18 (Different)}
        \label{fig:diff_wa_rn18}
    \end{subfigure}
    \hfill
    \begin{subfigure}[b]{0.32\textwidth}
        \centering
        \includegraphics[width=\linewidth,trim=0 1ex 0 1ex,clip]{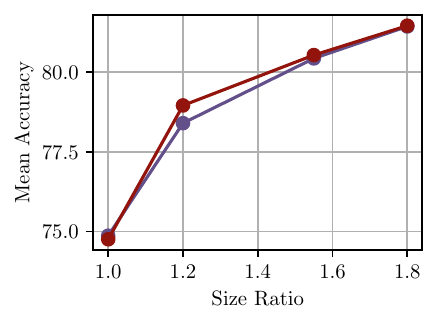}
        \caption{ResNet-50 (Different)}
        \label{fig:diff_wa_rn50}
    \end{subfigure}
    \hfill
    \begin{subfigure}[b]{0.32\textwidth}
        \centering
        \includegraphics[width=\linewidth,trim=0 1ex 0 1ex,clip]{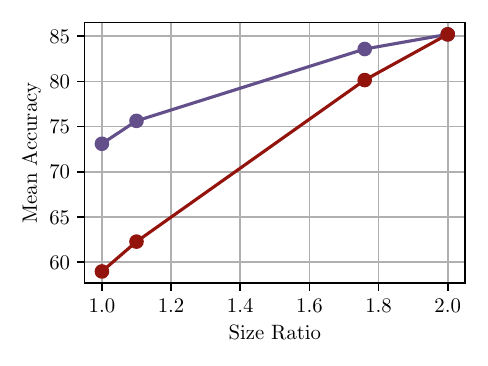}
        \caption{ResNet-101 (Different)}
        \label{fig:diff_wa_rn101}
    \end{subfigure}

    \caption{\textbf{Comparing \method{} and \methoda}: We find that \method{} is better for larger models while \methoda is better for smaller models.}\label{fig:weight_activation}
\end{figure*}
In \cref{fig:weight_activation}, we compare \methoda with \methodnospace, and find that \method{} is better for larger models, while at small sizes the two methods give a comparable performance.`
\section{Broader Impact}
Advances in model merging, especially through methods which do not require training data, can help further democratize machine learning by helping practitioners improve the capabilities of open source models. However, the risk of merged models inheriting biases of the component models still remains.

\subsection{Using synthetic data for merging}
\begin{figure*}
\begin{subfigure}[h]{0.29\textwidth}
    \centering
    \includegraphics[width=\linewidth]{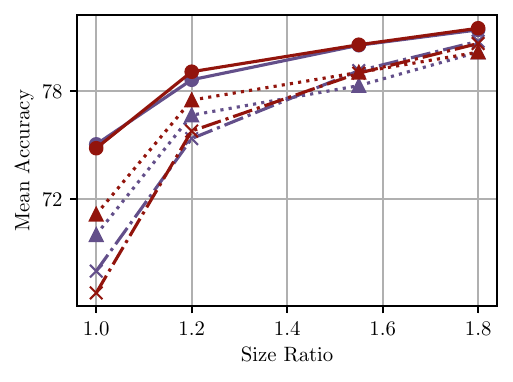}
    \vspace{-3.5ex}%
    \caption{Different Label Spaces}
    \label{fig:diff-label-space}
\end{subfigure}
\hfill
\begin{subfigure}[h]{0.29\textwidth}
    \centering
    \includegraphics[width=\linewidth]{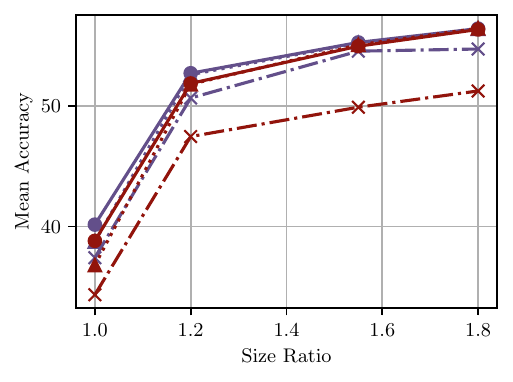}
    \vspace{-3.5ex}%
    \caption{Same Label Spaces}
    \label{fig:same-label-space}
\end{subfigure}
\hfill
\begin{subfigure}[h]{0.29\textwidth}
    \centering
    \includegraphics[width=\linewidth]{images/imagenet_data_ablations.pdf}
    \vspace{-3.5ex}%
    \caption{Imagenet}
    \label{fig:imagenet}
\end{subfigure}
\hfill
\begin{subfigure}[h]{0.1\textwidth}
    \centering
    \includegraphics[width=\linewidth]{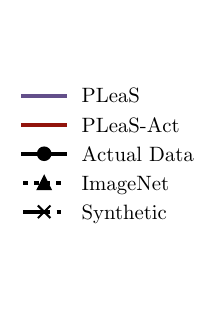}
\end{subfigure}
\vspace{-1.5ex}
\caption{\methodw with procedurally generated synthetic data. We find that using synthetic data is similar to using ImageNet}\label{fig:data-ablations}
\end{figure*}

We use synthetic images (e.g. procedurally generated data which can mimic the broad structure of real images) for merging. In \cref{fig:data-ablations}, we present the results of using such images with \methodwnospace.  This leads to a slight drop over using images from ImageNet, but the performance is close, and the gap closes as model size increases. 

\newpage

\end{document}